\pdfoutput=1

\documentclass[11pt,letterpaper]{article}
\usepackage[T1]{fontenc}
\usepackage{newtxtext}               %
\usepackage[margin=1in]{geometry}
\usepackage{graphicx}
\usepackage[hyphens]{url}
\usepackage{natbib}                  %
\usepackage{caption}
\usepackage{xcolor}
\usepackage{placeins}                %
\usepackage{amsmath,amssymb}
\usepackage{booktabs}
\usepackage{xspace}
\usepackage{enumitem}
\usepackage{multirow}   %
\usepackage{bm}         %
\usepackage{tikz}       %
\usetikzlibrary{arrows.meta,positioning,fit,backgrounds}
\definecolor{tbdblue}{rgb}{0.21,0.49,0.74}   %

\providecommand{\ie}{\emph{i.e.}\xspace}
\providecommand{\eg}{\emph{e.g.}\xspace}

\DeclareMathOperator{\Ex}{\mathbb{E}}
\newcommand{\real}{\mathbb{R}}
\newcommand{\setR}{\mathcal{R}}
\newcommand{\domega}{\Omega}

\newcommand{\doop}[1]{\mathrm{do}(#1)}

\newcommand{\method}{KCA\xspace}
\newcommand{\methodlong}{Kinematic Causal Attribution\xspace}
\newcommand{\Sgeo}{S_{\mathrm{geo}}}
\newcommand{\Smot}{S_{\mathrm{mot}}}
\newcommand{\Nec}{\mathcal{N}}
\newcommand{\Suf}{\mathcal{S}}
\newcommand{\abl}{\textsc{ablate}\xspace}
\newcommand{\frz}{\textsc{freeze}\xspace}
\newcommand{\dsp}{\textsc{displace}\xspace}
\newcommand{\ablgeo}{\textsc{abl-geo}\xspace}
\newcommand{\ace}{\mathrm{ACE}}
\newcommand{\PN}{\mathrm{PN}}
\newcommand{\PS}{\mathrm{PS}}

\usepackage{amsthm}
\theoremstyle{plain}
\newtheorem{proposition}{Proposition}
\theoremstyle{definition}

\newtheorem{assumption}{Assumption}

\usepackage{algorithm}
\usepackage{algpseudocode}



\newcommand{\tablecitesoff}{\renewcommand{\cite}[2][]{}}

\providecommand{\BBaccMBA}{88.2}
\providecommand{\BBaccPFT}{89.2}
\providecommand{\BBaccPST}{88.9}
\providecommand{\NumAblCtxBgDel}{0.126}

\providecommand{\NumAblCtxDel}{0.219}

\providecommand{\NumAblCtxFineDel}{0.212}

\providecommand{\NumAblCtxSsixtyfourDel}{0.231}

\providecommand{\NumAblLocalBgDel}{0.235}
\providecommand{\NumAblLocalBgEBPG}{0.49}

\providecommand{\NumAblLocalDel}{0.272}
\providecommand{\NumAblLocalEBPG}{0.54}
\providecommand{\NumAblLocalFineDel}{0.334}

\providecommand{\NumAblLocalMidShare}{0.68}

\providecommand{\NumAblLocalMotionFirstShare}{0.88}

\providecommand{\NumAblLocalShare}{0.73}
\providecommand{\NumCrossClips}{40}

\providecommand{\NumCtrlBgShape}{49}
\providecommand{\NumCtrlDspFlip}{90}
\providecommand{\NumCtrlFrzFlip}{0}
\providecommand{\NumCtrlMot}{1.00}
\providecommand{\NumCtrlMotCtx}{1.00}
\providecommand{\NumCtrlProxyDint}{0.02}
\providecommand{\NumCtrlShape}{0.00}
\providecommand{\NumCtrlShapeCtx}{0.00}

\providecommand{\NumDspFreezeFlip}{42}

\providecommand{\NumDspReverseFlip}{22}
\providecommand{\NumGTFitMedian}{95}
\providecommand{\NumGTFitPten}{92}
\providecommand{\NumGTMissing}{12}

\providecommand{\NumGTMovingPct}{13}
\providecommand{\NumGTStaticPct}{65}
\providecommand{\NumGTValid}{275}

\providecommand{\NumMBAKcaCtxMotShareAll}{0.75}

\providecommand{\NumPFTClips}{287}
\providecommand{\NumPFTClipsFive}{100}

\providecommand{\NumPFTExtremalDel}{0.634}

\providecommand{\NumPFTIGDel}{0.282}
\providecommand{\NumPFTIGEBPG}{0.55}

\providecommand{\NumPFTIGTime}{1.9}

\providecommand{\NumPFTKcaCtxDel}{0.179}

\providecommand{\NumPFTKcaCtxMotShareAll}{0.69}
\providecommand{\NumPFTKcaCtxMotShareMoving}{0.79}

\providecommand{\NumPFTKcaCtxSmotEBPG}{0.50}

\providecommand{\NumPFTKcaCtxTime}{2.3}
\providecommand{\NumPFTKcaDel}{0.256}

\providecommand{\NumPFTKcaTime}{0.5}

\providecommand{\NumPFTMotionPriorEBPG}{0.27}

\providecommand{\NumPFTOccamDel}{0.853}

\providecommand{\NumPFTOccamTime}{1.5}

\providecommand{\NumPFTOcclDel}{0.233}
\providecommand{\NumPFTOcclEBPG}{0.40}

\providecommand{\NumPFTRandomIns}{0.920}

\providecommand{\NumPFTRolloutDel}{0.649}

\providecommand{\NumPFTSTAADel}{0.542}

\providecommand{\NumPFTTAttrDel}{0.601}

\providecommand{\NumPKcaCtxIGDel}{{<}10^{-3}}

\providecommand{\NumPKcaCtxIGEBPG}{{<}10^{-3}}
\providecommand{\NumPKcaCtxOcclDel}{{<}10^{-3}}
\providecommand{\NumPKcaCtxOcclDelFive}{{=}0.10}

\providecommand{\NumPKcaOcclDel}{{=}0.011}
\providecommand{\NumPMBAKcaCtxIGDel}{{=}0.064}
\providecommand{\NumPMBAKcaCtxIGEBPG}{{=}0.73}
\providecommand{\NumPMBAKcaCtxOcclDel}{{=}0.83}

\providecommand{\NumPPSTKcaCtxIGDel}{{=}0.007}
\providecommand{\NumPPSTKcaCtxIGEBPG}{{=}0.63}
\providecommand{\NumPPSTKcaCtxOcclDel}{{=}0.38}

\providecommand{\NumSanGxI}{0.81}
\providecommand{\NumSanHeadKca}{0.12}
\providecommand{\NumSanKca}{$-$0.10}
\providecommand{\NumStabClips}{40}

\providecommand{\NumStabIGPart}{0.94}
\providecommand{\NumStabIGPoint}{0.92}

\providecommand{\NumStabKcaCtxAvgPart}{0.89}
\providecommand{\NumStabKcaCtxAvgPoint}{0.76}
\providecommand{\NumStabKcaCtxPart}{0.82}
\providecommand{\NumStabKcaCtxPoint}{0.62}
\providecommand{\NumStabKcaCtxShareSD}{0.04}

\providecommand{\NumStabOcclPoint}{0.71}
\providecommand{\NumStabPredSame}{99}

\providecommand{\NumStabSamplings}{5}
\providecommand{\NumTauPFTMBA}{0.71}
\providecommand{\NumTauPSTMBA}{0.81}
\providecommand{\NumTauPSTPFT}{0.71}
\providecommand{\NumXEval}{40}

\providecommand{\NumXGapHi}{0.32}
\providecommand{\NumXGapLo}{0.14}
\providecommand{\NumXGeoAcc}{83.5}

\providecommand{\NumXGeoHi}{0.67}
\providecommand{\NumXGeoLo}{0.53}

\providecommand{\NumXGeoPriorEBPG}{0.28}
\providecommand{\NumXGeoShare}{0.60}
\providecommand{\NumXGeoShareCtx}{0.60}
\providecommand{\NumXGeoShareSigned}{0.60}

\providecommand{\NumXMbaGapHi}{0.41}
\providecommand{\NumXMbaGapLo}{0.22}
\providecommand{\NumXMbaGeoAcc}{89.2}

\providecommand{\NumXMbaGeoHi}{0.69}
\providecommand{\NumXMbaGeoLo}{0.52}

\providecommand{\NumXMbaGeoShare}{0.61}
\providecommand{\NumXMbaGeoShareCtx}{0.58}
\providecommand{\NumXMbaGeoShareSigned}{0.61}
\providecommand{\NumXMbaMotAcc}{77.5}

\providecommand{\NumXMbaMotHi}{0.95}
\providecommand{\NumXMbaMotLo}{0.89}

\providecommand{\NumXMbaMotShare}{0.92}
\providecommand{\NumXMbaMotShareCtx}{0.90}
\providecommand{\NumXMbaMotShareSigned}{1.03}
\providecommand{\NumXMotAcc}{80.5}

\providecommand{\NumXMotHi}{0.88}
\providecommand{\NumXMotLo}{0.77}

\providecommand{\NumXMotPriorEBPG}{0.28}
\providecommand{\NumXMotShare}{0.83}
\providecommand{\NumXMotShareCtx}{0.80}
\providecommand{\NumXMotShareSigned}{0.84}

\providecommand{\NumXPstGapHi}{0.65}
\providecommand{\NumXPstGapLo}{0.48}
\providecommand{\NumXPstGeoAcc}{90.2}

\providecommand{\NumXPstGeoHi}{0.37}
\providecommand{\NumXPstGeoLo}{0.21}
\providecommand{\NumXPstGeoPriorEBPG}{0.28}
\providecommand{\NumXPstGeoShare}{0.30}
\providecommand{\NumXPstGeoShareCtx}{0.26}
\providecommand{\NumXPstGeoShareSigned}{0.16}
\providecommand{\NumXPstMotAcc}{80.5}

\providecommand{\NumXPstMotHi}{0.89}
\providecommand{\NumXPstMotLo}{0.82}
\providecommand{\NumXPstMotPriorEBPG}{0.28}
\providecommand{\NumXPstMotShare}{0.86}
\providecommand{\NumXPstMotShareCtx}{0.88}
\providecommand{\NumXPstMotShareSigned}{1.08}

\providecommand{\NumXSeedsCIExcl}{2}
\providecommand{\NumXSeedsSameDir}{3}
\providecommand{\NumXSeedsTotal}{6}
\providecommand{\NumXSeedsTrained}{3}
\providecommand{\NumXTest}{200}

\providecommand{\NumXThickShare}{0.88}
\providecommand{\NumXThickShareSigned}{0.49}
\providecommand{\NumXThinAcc}{50}
\providecommand{\NumXTrain}{163}

\usepackage[colorlinks,linkcolor=tbdblue,citecolor=tbdblue,urlcolor=tbdblue]{hyperref}
\let\cite\citep
\usepackage{etoolbox}
\AtBeginEnvironment{figure}{\setlength{\linewidth}{0.6\textwidth}\captionsetup{width=\linewidth}}

\newcommand{\orcid}[1]{\href{https://orcid.org/#1}{\raisebox{-0.15ex}{\tikz{%
  \fill[green!55!black!70!lime] (0,0) circle (0.8ex);
  \node[white,font=\fontsize{5.5}{5.5}\selectfont\sffamily\bfseries,inner sep=0pt] at (0,0) {iD};}}}}

\title{Kinematic Causal Attribution: Black-Box Explanations of\\Motion versus Geometry for 4D Point-Cloud Video Models}
\author{Zerui Wang\,\orcid{0000-0002-0403-3112}}
\hypersetup{pdfauthor={Zerui Wang},
  pdftitle={Kinematic Causal Attribution: Black-Box Explanations of Motion versus Geometry for 4D Point-Cloud Video Models}}

\date{}

\begin{document}
\maketitle
\begin{abstract}
4D perception models, which classify 3D point clouds that evolve over time, are being built into robots, assistive monitoring and AR, yet before deploying one a practitioner cannot check \emph{which cue} it relies on: how a body part moves, or what shape it has. Existing explanations do not fit this setting. 3D attribution is single-frame, video attribution is confined to RGB pixels, gradient maps need white-box access, attention is absent from state-space backbones, and point-level perturbations assume a point correspondence across frames that real depth and LiDAR data lack. We present \methodlong (\method), a black-box, interventional attribution method that queries only the model's outputs. Its \frz operator removes a region's motion while keeping its shape, and a shape-free proxy removes the shape while keeping the motion. Together they span a motion$\times$geometry cube for every tracked region, whose order-symmetric Shapley decomposition is complete and whose interaction is measured rather than assumed away. The operators rewrite per-frame point \emph{sets}, so they need no correspondences, and a context-averaged estimator keeps the scores reliable when a structure spans several regions. We validate \method on the official PSTNet, P4Transformer and Mamba4D trained on MSR-Action3D. The decisive test is an audit: two binary tasks are built from real clips whose classes differ only in the motion or only in the shape of the same arm, and the motion share of the arm's importance follows the defining cue on all three backbones (\NumXMotShare\ vs.\ \NumXGeoShare\ on P4Transformer, \NumXPstMotShare\ vs.\ \NumXPstGeoShare\ on PSTNet, \NumXMbaMotShare\ vs.\ \NumXMbaGeoShare\ on Mamba4D), a distinction no single attribution map can make. The split costs no faithfulness: the necessity maps have the lowest deletion AUC on P4Transformer and PSTNet, tie with the best baseline on Mamba4D, and lose their correlation with the trained model's once the weights are randomised. Skeleton ground truth, re-sampling stability and a cost study complete the validation.
\end{abstract}

\begin{figure*}[t]
\centering
\includegraphics[width=\textwidth]{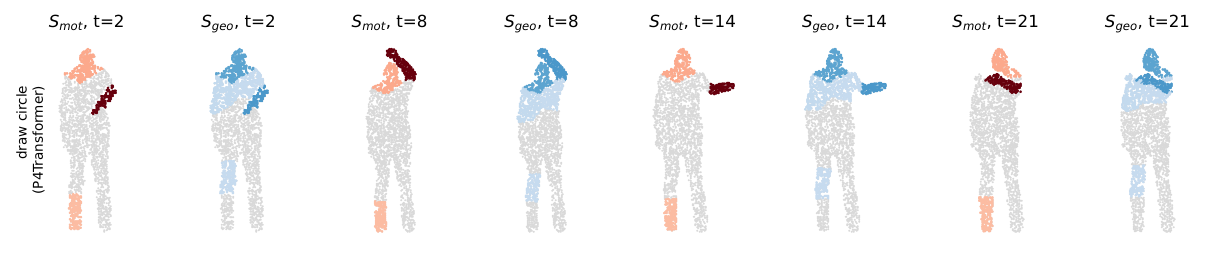}
\caption{\textbf{\methodlong\ (\method)} attributes a 4D model's decision to \emph{how} each spacetime region moves ($\Smot$, red) and to \emph{what shape} it has ($\Sgeo$, blue). It intervenes on the input and reads only the model's output. The clip is a real MSR-Action3D test clip (\emph{draw circle}) at four instants, explained for the official P4Transformer. The motion term is strongest on the drawing hand (weaker on the head and one foot), while geometry importance covers the upper body. Fig.~\ref{fig:qual} in the appendix shows two more actions on all three backbones, including one where the motion term falls on the torso. No attention, gradient or point correspondence is used, and the same code explains convolutional and attention-free state-space backbones (Sec.~\ref{sec:exp-cross}).}
\label{fig:teaser}
\end{figure*}

\section{Introduction}
\label{sec:intro}

Point-cloud video is the stream of 3D point sets produced by depth and LiDAR sensors, and it underpins action recognition, robotic manipulation, assistive monitoring and augmented reality. Dedicated \emph{4D} backbones now model it directly: point spatio-temporal convolutions~\cite{fan2021pstnet}, spatio-temporal transformers~\cite{fan2021p4transformer} and attention-free state-space models such as Mamba4D~\cite{liu2025mamba4d}. They are accurate but opaque. Before such a model is deployed, a practitioner should be able to check \emph{what} it bases a decision on. Did it label a clip ``high arm wave'' because of \emph{how} the arm moved, or because of \emph{what shape} the body had? A model that relies on shape where the task is defined by motion has learned a shortcut, as RGB action models do with static appearance~\cite{li2023staticbias}, and that shortcut can fail under new subjects, clothing or sensor placement.

\paragraph{What a deployable 4D explanation needs.}
We argue that an explanation method for deployed 4D models has to meet four requirements, and that no existing method meets them all.
\emph{(R1) Black-box access.} Deployed models are often compiled, quantised or served behind an interface, and the backbones differ in kind. Gradients need white-box access. Attention is a contested proxy for importance~\cite{jain2019attention,abnar2020rollout}, and Mamba4D has no attention to read. An explanation should therefore need only forward queries.
\emph{(R2) Real sensor data.} Sensors re-sample the surface in every frame, so ``the same point at two times'' does not exist. Any operator that moves or freezes indexed points silently breaks on real data.
\emph{(R3) The right question.} A prediction can depend on the \emph{shape} of a structure or on its \emph{motion}. In pixel space the two are entangled in one saliency value. In 4D they are physically distinct, yet 3D attribution is single-frame~\cite{zheng2019pointcloudsaliency,schinagl2022occam}, video attribution works on RGB pixels~\cite{price2020playfair,lee2025dance}, and no method attributes a decision to motion versus geometry.
\emph{(R4) Verifiable and affordable.} An explanation that will be acted on must be shown to be faithful to the model, must depend on the learned weights~\cite{adebayo2018sanity}, should be checked against ground truth where it exists, and must run at a practical cost.

\paragraph{Our approach.}
\methodlong (\method) treats explanation as intervention on tracked spacetime regions. Its key operator, \frz, holds a region static in its anchor-frame configuration: the region loses its motion and keeps its shape. A shape-free proxy that follows the region's centroid does the converse. The four resulting inputs form a motion$\times$geometry cube for each region. Averaging the cube's two paths yields an order-symmetric geometry term $\Sgeo$ and motion term $\Smot$ that sum exactly to the region's necessity $\Nec$, and the cube's mixed difference measures their interaction. All operators rewrite only a region's per-frame point \emph{sets}, so they need no correspondences. A context-averaged estimator, a Banzhaf-type average over random interventions on the other regions, keeps the scores reliable when a structure spans several regions. \method only queries the model, so it explains convolutional, attention and state-space backbones with the same code (Fig.~\ref{fig:teaser}).

\paragraph{Contributions.}
\begin{itemize}[leftmargin=1.2em,itemsep=2pt,topsep=2pt]
\item \textbf{\method, a black-box explanation that separates motion from geometry} (R1, R3). To our knowledge it is the first method that attributes a 4D model's decision to the \emph{motion} versus the \emph{geometry} of each spacetime region. Its decomposition is complete and order-symmetric (Shapley), and it reports the interaction instead of assuming it away (Sec.~\ref{sec:method}).
\item \textbf{A correspondence-free instantiation with guarantees} (R2). Set-based operators, a local and a context-averaged estimator, and the conditions under which each score is a causal effect.
\item \textbf{A validation protocol for deployment} (R4). On three unmodified official backbones we test faithfulness against 13 baselines, localisation against skeleton ground truth, dependence on the learned weights, stability under sensor re-sampling, and cost. We also give a \emph{crossover audit}: two tasks built from real clips whose classes differ only in the motion or only in the shape of one arm, which checks whether the explanation reports the cue a model was trained to use. The crossover is our primary evidence that the split is correct; the other tests show that it costs no faithfulness (Sec.~\ref{sec:exp}).
\end{itemize}

\section{Related Work}
\label{sec:related}

\paragraph{Point-cloud video models.}
Static point-cloud networks~\cite{qi2017pointnet,qi2017pointnetpp,wang2019dgcnn,wu2024ptv3} were extended to point-cloud video by MeteorNet~\cite{liu2019meteornet}, point spatio-temporal convolutions (PSTNet~\cite{fan2021pstnet}), point 4D convolution with self-attention (P4Transformer~\cite{fan2021p4transformer}; PST-Transformer~\cite{fan2023psttransformer}), primitive-level modelling~\cite{wen2022pptr}, temporal point patches~\cite{benshabat2024_3dinaction} and the attention-free state-space Mamba4D~\cite{liu2025mamba4d}. None of them explains its decisions. The ``motion saliency'' of recent pre-training~\cite{wang2026mosaic} selects tokens to mask, and Mamba4D's ``disentangled'' and ``causal'' refer to efficiency and scan order, not to attribution.

\paragraph{Attribution for images, video and point clouds.}
Surrogates~\cite{ribeiro2016lime,lundberg2017shap} gave way to occlusion and mask optimisation~\cite{petsiuk2018rise,fong2019extremal}, gradients~\cite{selvaraju2017gradcam,sundararajan2017ig} and transformer-specific methods~\cite{abnar2020rollout,chefer2021transformer}; attention as explanation is contested~\cite{jain2019attention} and unavailable for state-space models. Video methods attribute frames~\cite{price2020playfair}, read spatio-temporal attention~\cite{wang2025staa}, learn temporal concept bottlenecks~\cite{knab2025motif,lee2025dance} or test concept sensitivity~\cite{videotcav2024}. The closest in spirit, DANCE~\cite{lee2025dance}, separates motion, object and scene \emph{concepts}, but it is ante hoc, works in a learned concept space and is correlational; \method intervenes post hoc on physical 3D regions of any trained model. All of these work on 2D RGB. Point-cloud saliency~\cite{zheng2019pointcloudsaliency}, gradient maps~\cite{tayyub2022ape}, occlusion maps for LiDAR detectors~\cite{schinagl2022occam} and interpretable classifiers~\cite{devries2025pointmil} explain a \emph{single} cloud with no time axis. Interpretability for dynamic scene \emph{representations}---language-embedded 4D Gaussians~\cite{fiebelman2025_4legs,li2025_4dlangsplat}, static/dynamic separation for editing~\cite{kwon2025instruct4dgs}, importance for pruning~\cite{zhang2024lp3dgs}, prototypes for static splats~\cite{galus2026xsplain}---serves grounding, editing or compression, not the attribution of a recognition decision.

\paragraph{Causal attribution and trustworthy evaluation.}
LENS~\cite{watson2021lens} operationalises Pearl's necessity and sufficiency for tabular and text data, CaCE~\cite{goyal2019cace} contrasts interventional effects with concept sensitivity, and LD-ViCE~\cite{ldvice2025} generates RGB video counterfactuals; ``motion attribution'' has been used for \emph{training-data} attribution in video generation (Motive~\cite{wu2026motive}). None of them reaches 3D geometry and motion, and none separates the two. Uncertainty methods for 4D occupancy~\cite{reliocc2025,kalble2025evocc,occuq2025} ask how reliable a prediction is, not what caused it. Whether an explanation can be trusted is itself an empirical question. Deletion-based faithfulness~\cite{petsiuk2018rise,deyoung2020eraser} tests whether highly ranked inputs matter to the model, and model randomisation~\cite{adebayo2018sanity} tests whether the explanation depends on what the model learned. Point-cloud XAI still lacks an agreed evaluation standard~\cite{chakraborty2026benchmark}. We therefore validate \method with all three kinds of evidence: faithfulness, sanity checks and ground truth.

\section{Method}
\label{sec:method}

\begin{figure*}[t]
\centering
\resizebox{\textwidth}{!}{%
\begin{tikzpicture}[font=\small,>=Latex,node distance=6mm,
  box/.style={draw,rounded corners,align=center,minimum height=10mm,inner sep=3pt,fill=gray!8},
  op/.style={draw,rounded corners,align=center,minimum height=10mm,inner sep=3pt,fill=blue!8},
  res/.style={draw,rounded corners,align=center,minimum height=10mm,inner sep=3pt,fill=green!8}]
\node[box] (x) {Clip $X{=}\{P_t\}_{t=1}^{T}$\\[-1pt]\footnotesize no point identity};
\node[box,right=of x] (reg) {\textbf{(1) Tracked regions}\\[-1pt]\footnotesize per-frame sets $\setR_t$};
\node[op,right=of reg] (cube) {\textbf{(2) $(M,G)$ cube per region}\\[-1pt]\footnotesize $X$ \,$|$\, \frz \,$|$\, \ablgeo \,$|$\, \abl};
\node[op,right=of cube] (est) {\textbf{(3) Estimator}\\[-1pt]\footnotesize local $O(m)$ \,/\, context $O(S)$};
\node[res,right=of est] (scores) {\textbf{(4)} $\Smot,\Sgeo,\Delta_{\mathrm{int}}$\\[-1pt]\footnotesize $\Nec=\Smot+\Sgeo$};
\draw[->] (x)--(reg); \draw[->] (reg)--(cube); \draw[->] (cube)--(est); \draw[->] (est)--(scores);
\node[font=\footnotesize,above=0pt of est,text=blue!50!black] {model-agnostic ($f$ outputs only)};
\end{tikzpicture}}
\caption{\textbf{Method overview.} Regions are tracked through time as per-frame point \emph{sets}, so no cross-frame point correspondence is needed. Each region carries two factors---its geometry $G$ and its motion $M$---and four interventions realize the corners of the $(M,G)$ cube; every edge of the cube changes exactly one factor. A local or a context-averaged estimator turns the cube's outputs into order-symmetric motion and geometry importance whose interaction is measured. Only outputs of $f$ are used, so the same pipeline explains convolutional, attention and state-space backbones.}
\label{fig:method}
\end{figure*}
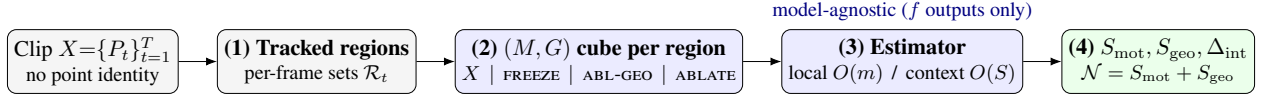

\method explains a trained 4D model by \emph{intervening} on spacetime regions of its input and measuring the causal effect on the prediction. After fixing the causal setting (Sec.~\ref{sec:method-prelim}) and notation (Sec.~\ref{sec:method-setup}), we define correspondence-free interventions (Sec.~\ref{sec:method-interventions}) and the causal scores they induce (Sec.~\ref{sec:method-causal}), derive the motion-vs-geometry decomposition that is our central contribution (Sec.~\ref{sec:method-decomp}), establish its properties (Sec.~\ref{sec:method-theory}), and give two estimators (Sec.~\ref{sec:method-eff}). Fig.~\ref{fig:method} gives an overview.

\subsection{Preliminaries: causal attribution}
\label{sec:method-prelim}
We adopt the interventional view of explanation~\cite{watson2021lens,janzing2020feature,pearl2009causality}: the importance of an input component is the change it \emph{causes} in the output under an intervention $\doop{\cdot}$, rather than a gradient or attention read-out. For a target $\varphi$ and a manipulable factor $Z$, the \emph{average causal effect} (ACE) of setting $Z{=}z$ against a reference $z_0$ is $\ace_Z = \Ex[\varphi\mid \doop{Z{=}z}]-\Ex[\varphi\mid \doop{Z{=}z_0}]$. Pearl's probability of \emph{necessity} ($\PN$) asks whether removing a cause flips the outcome, and of \emph{sufficiency} ($\PS$) whether the cause alone produces it~\cite{pearl2009causality}; \citet{watson2021lens} operationalize local $\PN/\PS$ for feature subsets. We (i) instantiate these for \emph{spacetime} regions of a 4D model, without assuming point correspondences, and (ii) factor each region's effect into \emph{motion} and \emph{geometry}---a factorization unique to dynamic-3D data. Appendix~\ref{supp:notation} summarises the notation and gives pseudocode (Alg.~\ref{alg:kca}).

\subsection{Problem formulation}
\label{sec:method-setup}
A point-cloud video is a sequence $X=\{P_t\}_{t=1}^{T}$, $P_t=\{\mathbf{p}_{t,i}\}_{i=1}^{N_t}\subset\real^{3}$. Depth and LiDAR sensors re-sample the surface in every frame, so there is \emph{no point identity} across frames: index $i$ in frame $t$ and index $i$ in frame $t'$ are unrelated points. A trained model $f$ maps $X$ to logits; we explain $\varphi(X)=f(X)_c$ for the predicted class $c$ (any scalar read-out is admissible, including non-differentiable ones). The spacetime support is partitioned into $m$ \emph{tracked regions}: region $\setR$ is a sequence of per-frame point sets $\{\setR_t\subseteq P_t\}_t$ covering the same physical structure (tracked k-means supervoxels; Appendix~\ref{supp:impl}). \method assigns each region $\Sgeo(\setR)$ and $\Smot(\setR)$, aggregated into per-point fields.

\subsection{Correspondence-free spacetime interventions}
\label{sec:method-interventions}
A region has two factors: its \emph{geometry} $G$ (the shape it has in each frame) and its \emph{motion} $M$ (how it moves through time). We realize every combination with operators that act only on the per-frame sets $\setR_t$---never on indexed points:
\begin{itemize}[leftmargin=1.2em,itemsep=1pt,topsep=2pt]
\item $\doop{\frz,\setR}$ \,($M{=}0$, $G{=}1$): every $\setR_t$ is replaced by the region's anchor-frame set $\setR_{t_0}$ (re-sampled to $|\setR_t|$ points): the region stays static in its $t_0$ configuration, with its real shape. This removes \emph{all} of its motion---translation, rotation and articulation alike---without any correspondence.
\item $\doop{\ablgeo,\setR}$ \,($M{=}1$, $G{=}0$): $\setR_t$ is replaced by a \emph{shape-free proxy}---a point mass---that follows the region's centroid trajectory $c_t(\setR)$: the region keeps its motion but loses its shape.
\item $\doop{\abl,\setR}$ \,($M{=}0$, $G{=}0$): the same proxy, held static at $c_{t_0}(\setR)$.
\end{itemize}
The four corners $X$, $\doop{\frz,\setR}X$, $\doop{\ablgeo,\setR}X$, $\doop{\abl,\setR}X$ form the region's $(M,G)$ cube, and \emph{every edge changes exactly one factor}. This is the property that makes the decomposition below meaningful: had \abl\ \emph{deleted} the region while \ablgeo\ kept a proxy, the edge $\ablgeo\!\to\!\abl$ would change presence as well as motion, and a static region's presence would be booked as ``motion'' (we measure this failure mode in Sec.~\ref{sec:exp-ablation}). Keeping the proxy at the region's own location also keeps every corner on the scene's support. A third operator, $\doop{\dsp,\setR}$, replays the region's sets in reversed or permuted order ($\setR_t\!\leftarrow\!\setR_{\pi(t)}$): a sign-sensitive probe of motion \emph{direction}. When point identity does exist (synthetic data), all operators reduce exactly to their index-aligned forms, \eg $\mathbf p_{t,i}\!\leftarrow\!\mathbf p_{t_0,i}$ for \frz.

\subsection{Causal attribution: necessity and sufficiency}
\label{sec:method-causal}
\emph{Necessity} measures the drop when a region's spatio-temporal content---shape and motion---is replaced by the static proxy,
\begin{equation}
\Nec(\setR)\;=\;\varphi(X)\;-\;\varphi\!\big(\doop{\abl,\setR}\,X\big),
\label{eq:necessity}
\end{equation}
and \emph{sufficiency} how much the region alone preserves, $\Suf(\setR)=\varphi(\doop{\abl,\domega\!\setminus\!\setR}X)-\varphi(\doop{\abl,\domega}X)$. High $\Nec$ marks regions the decision \emph{depends on}. Unlike attention or gradients, both are defined for \emph{any} $f$, including attention-free state-space models.

\subsection{Disentangling motion from geometry}
\label{sec:method-decomp}
Necessity conflates a region's motion and its geometry; the cube separates them. Writing $\varphi\!\equiv\!\varphi(X)$ and $\varphi_{\mathrm{op}}\!\equiv\!\varphi(\doop{\mathrm{op},\setR}X)$, the two paths from $X$ to $\doop{\abl,\setR}X$ remove motion first or geometry first; averaging them yields the \emph{order-symmetric} scores
\begin{align}
\Smot(\setR)&=\tfrac12\big[(\varphi-\varphi_{\frz})+(\varphi_{\ablgeo}-\varphi_{\abl})\big],\nonumber\\
\Sgeo(\setR)&=\tfrac12\big[(\varphi-\varphi_{\ablgeo})+(\varphi_{\frz}-\varphi_{\abl})\big],
\label{eq:sym}
\end{align}
each an average of two single-factor contrasts. $\Smot$ asks \emph{does the decision depend on how this structure moves?}, $\Sgeo$ \emph{does it depend on what shape it has?}---a question no prior attribution can pose. Their gap to the single-order scores is the interaction
\begin{equation}
\Delta_{\mathrm{int}}(\setR)=\varphi-\varphi_{\frz}-\varphi_{\ablgeo}+\varphi_{\abl},
\label{eq:dint}
\end{equation}
which we \emph{measure} rather than assume away.

\subsection{Theoretical properties}
\label{sec:method-theory}
\method queries $f$ only through outputs of intervened inputs, so its guarantees hold for every architecture. The headline results require only interventional validity \textbf{(I)}; separability \textbf{(S)} is \emph{measured}, not assumed (Tab.~\ref{tab:assump}; proofs in Appendix~\ref{supp:proofs}).

\begin{assumption}[Interventional validity \textbf{(I)} \& separability \textbf{(S)}]
\label{asm:factor}
A region's effect on $\varphi$ is mediated by a geometry factor $G_\setR$ and a motion factor $M_\setR$. \textbf{(I)} \frz, \ablgeo\ and \abl\ set the targeted factor(s) without altering off-region content (they only rewrite $\setR_t$). \textbf{(S)} \emph{(diagnostic)} $\varphi$ is additively separable in $(M_\setR,G_\setR)$, \ie the discrete mixed difference vanishes.
\end{assumption}

\begin{proposition}[Completeness and the order-2 Shapley decomposition]
\label{prop:interact}
For any $f$, $\Smot(\setR)+\Sgeo(\setR)=\Nec(\setR)$. $(\Smot,\Sgeo)$ is the Shapley value of the two-player game $\{M_\setR,G_\setR\}$ with worth $v(\{\cdot\})=\varphi$ at the corresponding corner; $\Delta_{\mathrm{int}}$ is its Shapley interaction index~\cite{grabisch1999interaction}; each single-order score deviates from the symmetric one by exactly $\pm\Delta_{\mathrm{int}}/2$; and $\Delta_{\mathrm{int}}{=}0$ iff \textbf{(S)} holds for $\setR$.
\end{proposition}

\begin{proposition}[Causal identification of motion]
\label{prop:ace}
Under \textbf{(I)}, $\varphi-\varphi_{\frz}$ is the interventional effect of region $\setR$'s motion at its real geometry and $\varphi_{\ablgeo}-\varphi_{\abl}$ the same effect for a shape-free body (CaCE-type contrasts~\cite{goyal2019cace}); $\Smot$ is their average. Under \textbf{(S)} the two coincide and equal the marginal motion ACE; symmetrically for $\Sgeo$.
\end{proposition}

\begin{proposition}[Necessity correspondence]
\label{prop:pnps}
With the decision event $E{=}\mathbf 1[\varphi{\ge}\kappa]$, $\Nec(\setR)$ is the margin relaxation of the empirical probability of necessity $\PN(\setR)=\Pr[E(\doop{\abl,\setR}X){=}0\mid E(X){=}1]$~\cite{pearl2009causality,watson2021lens}; under monotonicity and exogeneity \textbf{(M)} it is identified by the excess-risk ratio. $\Suf$ is the dual $\PS$ score.
\end{proposition}

\subsection{Estimation}
\label{sec:method-eff}
\paragraph{Local estimator.} Evaluating the cube of every region at the observed input costs $3m{+}1$ forward passes, which we batch. The local scores answer ``what happens if this region alone is intervened on''---the classic occlusion question.

\paragraph{Context-averaged estimator.} Local contrasts under-credit \emph{redundant} structure: if a decision-relevant part spans several regions, removing any one of them leaves enough of the others, and every local necessity is small. We therefore also average the same contrasts over random \emph{contexts} in which the other regions are themselves intervened. We draw $S$ joint states with every factor $M_{\setR},G_{\setR}\sim\mathrm{Bernoulli}(q)$ independently, evaluate $\varphi$ once per state (all regions intervened jointly), and fit
\begin{equation}
\varphi\approx\beta_0+\textstyle\sum_{\setR}\big(\mu_{\setR}M_{\setR}+\gamma_{\setR}G_{\setR}+\iota_{\setR}M_{\setR}G_{\setR}\big)
\label{eq:context}
\end{equation}
by least squares, then set $\Smot{=}\mu+\iota/2$, $\Sgeo{=}\gamma+\iota/2$, $\Delta_{\mathrm{int}}{=}\iota$, $\Nec{=}\mu{+}\gamma{+}\iota$.
\begin{proposition}[Context-averaged decomposition]
\label{prop:context}
Eq.~\eqref{eq:context} satisfies completeness $\Smot+\Sgeo=\Nec$ exactly. If $\varphi$ has no cross-region interactions, its population solution equals Eq.~\eqref{eq:sym} (for any context); otherwise it is the context-averaged (Banzhaf-type~\cite{banzhaf1965}) version of Eq.~\eqref{eq:sym} under the sampling distribution.
\end{proposition}
The cost is $S$ passes, independent of $m$. Both estimators use the same interventions; the local one is exact for a single context and cheaper, the context one is robust to redundancy.

\subsection{Model-agnostic instantiation}
\label{sec:method-inst}
We explain three architecturally distinct official backbones with the \emph{same} code: the convolutional PSTNet~\cite{fan2021pstnet}, the attention-based P4Transformer~\cite{fan2021p4transformer}, and the attention-free state-space Mamba4D~\cite{liu2025mamba4d}. The interventions and scores (Eqs.~\eqref{eq:necessity}--\eqref{eq:context}) touch the model only through its outputs; the only backbone-specific code is the input contract (a fixed number of points per frame, so an intervened frame is re-sampled from its remaining points, as the official data loaders do). Attention is never used: \method is equally defined on models that have none.

\section{Experiments}
\label{sec:exp}
We ask four questions. \textbf{Q1} Is \method's necessity map faithful? \textbf{Q2} Does one code path explain convolutional, attention and state-space models? \textbf{Q3} Is the motion/geometry split \emph{correct}? \textbf{Q4} What do its components contribute and cost?

\subsection{Setup}
\label{sec:exp-setup}
\paragraph{Data and ground truth.}
We explain action recognition on MSR-Action3D~\cite{li2010msraction3d}: 567 depth videos of 20 actions by 10 subjects, with the standard split (subjects 1--5 for training, 270 videos; 6--10 for testing, 297 videos). Depth maps become point-cloud videos with the MeteorNet protocol~\cite{liu2019meteornet} used by all three official backbones; we use centred 24-frame clips with 2048 points per frame, re-sampled independently in every frame (no point identity). MSR-Action3D also ships 20-joint skeletons. We lift them into the same coordinates, where a median \NumGTFitMedian\% of the points lie within about 12\,cm of a bone, and label every point with its nearest bone. Some videos store no skeleton for a run of frames; every ground-truth-based number leaves out the \NumGTMissing\ benchmark clips concerned. Points of bones that travel at least half as far as the clip's most active bone form the \emph{moving} ground truth (on average \NumGTMovingPct\% of a clip's points); points of bones that travel at most a fifth as far form the \emph{static} ground truth (\NumGTStaticPct\%; Appendix~\ref{supp:impl}). The main benchmark explains every test video that is at least 24 frames long (\NumPFTClips\ clips; \NumGTValid\ with a complete skeleton). The cross-backbone study, the ablations and the stability study use the first two test clips of every action (\NumCrossClips\ clips).

\paragraph{Backbones.}
We explain the \emph{unmodified official code} of the convolutional PSTNet~\cite{fan2021pstnet}, the attention-based P4Transformer~\cite{fan2021p4transformer} and the attention-free state-space Mamba4D~\cite{liu2025mamba4d}. No pretrained weights exist, so we train each from scratch at its official width on one A100 GPU, replacing the CUDA-only operators by numerically equivalent PyTorch versions (Appendix~\ref{supp:impl}). Test accuracies (video level) are \BBaccPFT/\BBaccPST/\BBaccMBA\,\% for P4Transformer/PSTNet/Mamba4D (published: 90.9/91.2/92.7). A second, laptop-scale training (reduced width, 1024 points) supports the same conclusions; Appendix~\ref{supp:scale} compares the two.

\paragraph{Baselines.}
\emph{Floors:} random; geometry-edge (local point spacing); motion prior (model-blind nearest-neighbour flow between frames). \emph{Gradient:} Grad$\times$Input, SmoothGrad~\cite{smilkov2017smoothgrad}, Integrated Gradients~\cite{sundararajan2017ig}, and 3D Grad-CAM~\cite{selvaraju2017gradcam} on the tube-embedding tokens. \emph{Perturbation:} region occlusion (deleting each tracked region, as in point-cloud saliency~\cite{zheng2019pointcloudsaliency}), OccAM~\cite{schinagl2022occam}, and extremal perturbations~\cite{fong2019extremal}. \emph{Attention (P4Transformer only):} rollout~\cite{abnar2020rollout}, transformer attribution~\cite{chefer2021transformer}, and raw attention: the attention each token receives in the last layer, used \emph{as} the explanation.
\paragraph{\method.} $m{=}12$ tracked regions; local estimator (37 passes) and context estimator ($S{=}256$, $q{=}\tfrac12$). An attribution map gives each point its region's $\Nec$; localisation uses $\Smot$ (moving GT) and $\Sgeo$ (static GT).

\paragraph{Metrics.}
\emph{Faithfulness} (no GT): deletion/insertion AUC of the target-class probability~\cite{petsiuk2018rise}, comprehensiveness/sufficiency (AOPC over $\{1,5,10,20,50\}\%$)~\cite{deyoung2020eraser}, and Sensitivity-$n$~\cite{ancona2018sensitivity}; a deleted point is removed and its frame re-sampled. \emph{Localisation} (skeleton GT): Pointing Game~\cite{zhang2018pointing} and energy-based EBPG~\cite{wang2020scorecam}. \emph{Sanity:} cascading model randomisation~\cite{adebayo2018sanity}. \emph{Statistics.} Every score is computed per clip and reported as the mean over clips. A $p$-value compares two methods on the same clips with a two-sided paired Wilcoxon signed-rank test~\cite{wilcoxon1945}; $p$-values in the text are uncorrected, and a $^\ast$ in a table marks \method\ better than the strongest baseline after Bonferroni correction over that table's metrics~\cite{dunn1961}. Intervals are 95\% percentile bootstrap intervals over clips (2000 resamples)~\cite{efron1979bootstrap}. Appendix~\ref{supp:provenance} lists the data, models and code behind every table and figure.

\begin{table*}[t]
\centering\small\tablecitesoff
\begin{tabular}{l cccc}
\toprule
Method & Del.$\downarrow$ & Comp.$\uparrow$ & Ins.$\uparrow^\dagger$ & Suff.$\downarrow^\dagger$\\
\midrule
Random (floor) & 0.916{\scriptsize$\pm$0.14} & 0.003{\scriptsize$\pm$0.05} & 0.920{\scriptsize$\pm$0.14} & 0.041{\scriptsize$\pm$0.12}\\
Geometry-edge (floor) & 0.890{\scriptsize$\pm$0.19} & 0.011{\scriptsize$\pm$0.11} & 0.903{\scriptsize$\pm$0.16} & 0.103{\scriptsize$\pm$0.21}\\
Motion-prior (floor) & 0.505{\scriptsize$\pm$0.38} & 0.277{\scriptsize$\pm$0.33} & 0.860{\scriptsize$\pm$0.16} & 0.365{\scriptsize$\pm$0.22}\\
Grad$\times$Input & 0.655{\scriptsize$\pm$0.28} & 0.086{\scriptsize$\pm$0.18} & 0.722{\scriptsize$\pm$0.25} & 0.332{\scriptsize$\pm$0.28}\\
SmoothGrad~\cite{smilkov2017smoothgrad} & 0.772{\scriptsize$\pm$0.24} & 0.031{\scriptsize$\pm$0.12} & \underline{0.915{\scriptsize$\pm$0.14}} & \underline{0.083{\scriptsize$\pm$0.16}}\\
Integrated Grad.~\cite{sundararajan2017ig} & 0.282{\scriptsize$\pm$0.26} & 0.347{\scriptsize$\pm$0.24} & 0.862{\scriptsize$\pm$0.17} & 0.129{\scriptsize$\pm$0.19}\\
3D Grad-CAM~\cite{selvaraju2017gradcam} & 0.713{\scriptsize$\pm$0.26} & 0.051{\scriptsize$\pm$0.12} & 0.554{\scriptsize$\pm$0.30} & 0.659{\scriptsize$\pm$0.30}\\
Region occlusion~\cite{zheng2019pointcloudsaliency} & \underline{0.233{\scriptsize$\pm$0.23}} & \underline{0.416{\scriptsize$\pm$0.23}} & 0.784{\scriptsize$\pm$0.17} & 0.531{\scriptsize$\pm$0.24}\\
OccAM~\cite{schinagl2022occam} & 0.853{\scriptsize$\pm$0.24} & 0.059{\scriptsize$\pm$0.13} & \textbf{0.953{\scriptsize$\pm$0.07}} & \textbf{$-$0.004{\scriptsize$\pm$0.11}}\\
Extremal Pert.~\cite{fong2019extremal} & 0.634{\scriptsize$\pm$0.22} & 0.052{\scriptsize$\pm$0.10} & 0.309{\scriptsize$\pm$0.26} & 0.830{\scriptsize$\pm$0.25}\\
Attn.\ Rollout~\cite{abnar2020rollout} & 0.649{\scriptsize$\pm$0.27} & 0.073{\scriptsize$\pm$0.15} & 0.769{\scriptsize$\pm$0.24} & 0.431{\scriptsize$\pm$0.26}\\
Transformer Attr.~\cite{chefer2021transformer} & 0.601{\scriptsize$\pm$0.28} & 0.100{\scriptsize$\pm$0.17} & 0.766{\scriptsize$\pm$0.24} & 0.424{\scriptsize$\pm$0.28}\\
Raw attention (last layer) & 0.542{\scriptsize$\pm$0.28} & 0.127{\scriptsize$\pm$0.16} & 0.688{\scriptsize$\pm$0.25} & 0.551{\scriptsize$\pm$0.29}\\
\midrule
\textbf{\method} (local) & 0.256{\scriptsize$\pm$0.25} & 0.403{\scriptsize$\pm$0.24} & 0.793{\scriptsize$\pm$0.17} & 0.511{\scriptsize$\pm$0.23}\\
\textbf{\method} (context) & \textbf{0.179{\scriptsize$\pm$0.19}}$^\ast$ & \textbf{0.453{\scriptsize$\pm$0.21}}$^\ast$ & 0.761{\scriptsize$\pm$0.18} & 0.546{\scriptsize$\pm$0.24}\\
\bottomrule
\end{tabular}
\caption{\textbf{Faithfulness} on P4Transformer, MSR-Action3D test set (all \NumPFTClips\ test clips; mean$\pm$std over clips). Del.: deletion AUC of the target-class probability; Comp.: comprehensiveness. Both remove the top-ranked points. Ins.\ and Suff.\ ($\dagger$) \emph{keep} the top-ranked points, which rewards spatially scattered rankings on point clouds (a random subset preserves the whole body; see the random floor). \textbf{Bold}: best, \underline{underlined}: second, floors excluded. $^\ast$: \method\ significantly better than the strongest baseline after Bonferroni correction.}
\label{tab:faithfulness}
\end{table*}

\subsection{Faithfulness (Q1)}
\label{sec:exp-main}
\begin{figure}[t]
\centering
\includegraphics[width=\linewidth]{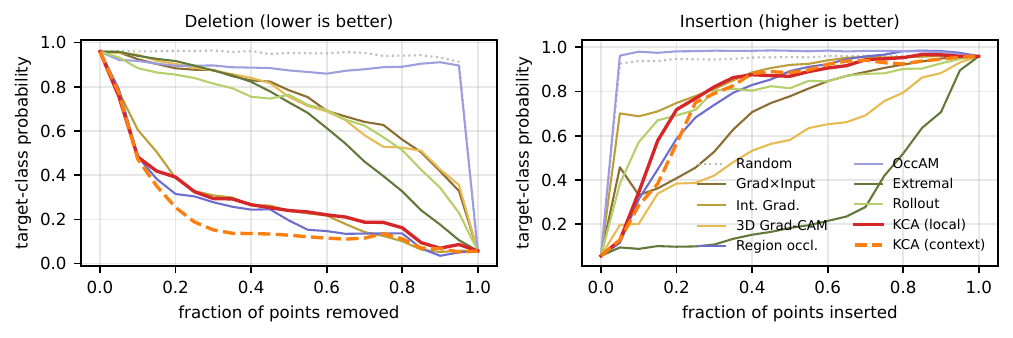}
\caption{\textbf{Deletion and insertion curves} on P4Transformer (mean over the \NumPFTClips\ clips of Tab.~\ref{tab:faithfulness}): target-class probability vs.\ fraction of points removed (left) or inserted (right), in 20 steps.}
\label{fig:curves}
\end{figure}

Tab.~\ref{tab:faithfulness} and Fig.~\ref{fig:curves} compare the necessity maps. \method's context estimator attains the lowest deletion AUC (\NumPFTKcaCtxDel; local estimator \NumPFTKcaDel) and the highest comprehensiveness. Its deletion AUC is significantly below that of region occlusion (\NumPFTOcclDel; paired Wilcoxon $p\NumPKcaCtxOcclDel$) and of Integrated Gradients (\NumPFTIGDel; $p\NumPKcaCtxIGDel$), and far ahead of attention rollout (\NumPFTRolloutDel), transformer attribution (\NumPFTTAttrDel) and raw attention (\NumPFTSTAADel): on the very backbone that has attention, reading it out is a poor explanation. Two findings deserve comment. \emph{(i) Region occlusion is \method's necessity with a deletion corner.} Replacing the static proxy of Sec.~\ref{sec:method-interventions} by deletion makes the local necessity map \emph{identical} to region occlusion, which the deletion metric (itself deletion-based) favours over the local estimator ($p\NumPKcaOcclDel$); the context estimator more than recovers the gap, and the proxy is what makes the motion/geometry split meaningful (Sec.~\ref{sec:exp-mpg}, Tab.~\ref{tab:ablation}). \emph{(ii) Point-cloud models shrug off random point dropout.} Deleting points in random order leaves the prediction almost intact (Fig.~\ref{fig:curves}), so point-level random masking (OccAM, \NumPFTOccamDel) and soft-mask optimisation (Extremal, \NumPFTExtremalDel) carry little signal at this budget, and \emph{keep}-based metrics reward scattered rankings: the random floor scores an insertion AUC of \NumPFTRandomIns. We therefore read faithfulness mainly from the deletion-based columns.

\begin{table*}[t]
\centering\small\tablecitesoff
\begin{tabular}{l cc cc cc}
\toprule
& \multicolumn{2}{c}{PSTNet} & \multicolumn{2}{c}{P4Transformer} & \multicolumn{2}{c}{Mamba4D}\\
\cmidrule(lr){2-3}\cmidrule(lr){4-5}\cmidrule(lr){6-7}
Method & Del.$\downarrow$ & EBPG$\uparrow$ & Del.$\downarrow$ & EBPG$\uparrow$ & Del.$\downarrow$ & EBPG$\uparrow$\\
\midrule
Random (floor) & 0.950 & 0.098 & 0.929 & 0.098 & 0.902 & 0.098\\
Integrated Grad.~\cite{sundararajan2017ig} & 0.303 & \textbf{0.390} & \underline{0.230} & \textbf{0.567} & 0.329 & \textbf{0.456}\\
Region occlusion~\cite{zheng2019pointcloudsaliency} & \underline{0.252} & 0.329 & \underline{0.230} & 0.394 & \textbf{0.246} & 0.351\\
OccAM~\cite{schinagl2022occam} & 0.845 & 0.098 & 0.865 & 0.098 & 0.778 & 0.098\\
Extremal Pert.~\cite{fong2019extremal} & 0.932 & 0.098 & 0.621 & 0.076 & 0.577 & 0.085\\
Attn.\ Rollout~\cite{abnar2020rollout} & \emph{n/a} & \emph{n/a} & 0.667 & 0.141 & \emph{n/a} & \emph{n/a}\\
Raw attention (last layer) & \emph{n/a} & \emph{n/a} & 0.591 & 0.201 & \emph{n/a} & \emph{n/a}\\
\midrule
\textbf{\method} (local) & 0.262 & 0.326 & 0.272 & 0.504 & 0.314 & 0.392\\
\textbf{\method} (context) & \textbf{0.217} & \underline{0.361} & \textbf{0.219} & \underline{0.506} & \underline{0.269} & \underline{0.433}\\
\bottomrule
\end{tabular}
\caption{\textbf{Across backbones} (the same \NumCrossClips\ test clips, two per action, for every backbone; means over clips): deletion AUC and localisation of the moving ground truth (EBPG, on the clips with a complete skeleton; $\Smot$ for \method). Attention methods are undefined on PSTNet and Mamba4D (n/a). Kendall's $\tau$~\cite{kendall1938} between the rankings by Del.\ of the 7 methods defined on all three backbones: PSTNet--P4Transf.\ \NumTauPSTPFT, PSTNet--Mamba4D \NumTauPSTMBA, P4Transf.--Mamba4D \NumTauPFTMBA.}
\label{tab:cross}
\end{table*}

\subsection{Across backbones (Q2)}
\label{sec:exp-cross}
Tab.~\ref{tab:cross} runs the same code on the three official architectures. \method is defined on all of them---the attention methods are not. On every backbone its context estimator has the lowest or second-lowest deletion AUC, statistically tied with region occlusion on PSTNet and Mamba4D (paired Wilcoxon $p\NumPPSTKcaCtxOcclDel$, $p\NumPMBAKcaCtxOcclDel$) and below Integrated Gradients, significantly on PSTNet ($p\NumPPSTKcaCtxIGDel$) but not on Mamba4D ($p\NumPMBAKcaCtxIGDel$); its motion term localises the moving parts on par with Integrated Gradients, the best single map ($p\NumPPSTKcaCtxIGEBPG$, $p\NumPMBAKcaCtxIGEBPG$), and at least as well as region occlusion. The method ranking is consistent across backbones (Kendall $\tau$, caption), and the state-space Mamba4D, which has no attention to read, is explained exactly like the others. On PSTNet and Mamba4D every informative map puts less mass on the moving parts than on P4Transformer: these models spread their evidence more widely (Sec.~\ref{sec:exp-qual}).

\begin{table}[t]
\centering\small\tablecitesoff\setlength{\tabcolsep}{3pt}
\begin{tabular}{l cc c cc}
\toprule
& \multicolumn{2}{c}{$\to$ moving GT} & $\to$ static & \multicolumn{2}{c}{crossover: cue =}\\
\cmidrule(lr){2-3}\cmidrule(lr){4-4}\cmidrule(lr){5-6}
Method & PG$\uparrow$ & EBPG$\uparrow$ & PG$\uparrow$ & motion & shape\\
\midrule
Grad$\times$Input & 0.46 & 0.35 & -- & -- & --\\
Integrated Grad.~\cite{sundararajan2017ig} & \textbf{0.92} & \textbf{0.55} & -- & -- & --\\
3D Grad-CAM~\cite{selvaraju2017gradcam} & 0.19 & 0.10 & -- & -- & --\\
Region occlusion~\cite{zheng2019pointcloudsaliency} & 0.40 & 0.40 & -- & -- & --\\
OccAM~\cite{schinagl2022occam} & 0.17 & 0.13 & -- & -- & --\\
Attn.\ Rollout~\cite{abnar2020rollout} & 0.19 & 0.17 & -- & -- & --\\
Motion-prior (floor) & 0.79 & 0.27 & -- & -- & --\\
\midrule
\textbf{\method} (local) & 0.47 & 0.51 & 0.41 & 0.83 & 0.60\\
\textbf{\method} (context) & 0.48 & 0.50 & 0.38 & 0.80 & 0.60\\
\bottomrule
\end{tabular}
\caption{\textbf{Motion vs.\ geometry} (P4Transformer). Left: localisation against the skeleton ground truth on the \NumGTValid\ clips of Tab.~\ref{tab:faithfulness} with a complete skeleton (PG: Pointing Game; $\Smot$ against moving, $\Sgeo$ against static parts). Right: \emph{crossover} (Sec.~\ref{sec:exp-mpg}, \NumXEval\ test clips per task), motion share $\Smot^+/(\Smot^++\Sgeo^+)$ of the positive importance on the arm; single maps have no split (--).}
\label{tab:mpg}
\end{table}

\subsection{Motion versus geometry (Q3)}
\label{sec:exp-mpg}
\paragraph{Localisation.} $\Smot$ places \NumPFTKcaCtxSmotEBPG\ of its positive mass on the skeleton's moving parts (EBPG, Tab.~\ref{tab:mpg}), above region occlusion (\NumPFTOcclEBPG) and far above the model-blind motion prior (\NumPFTMotionPriorEBPG), whose mass is spread over sensor jitter, but below the best single map, Integrated Gradients (\NumPFTIGEBPG; $p\NumPKcaCtxIGEBPG$). The single most-salient point of Integrated Gradients hits the moving parts more often (PG), since region-constant maps have no ranking \emph{within} a region. On moving parts, \NumPFTKcaCtxMotShareMoving\ of \method's positive importance is motion, against \NumPFTKcaCtxMotShareAll\ over the whole body. Localisation, however, is confounded: moving parts are also geometrically distinctive, so a map that highlights the arm for \emph{any} reason scores well.
\paragraph{Crossover: the decisive test.} We therefore build two binary tasks from real clips in which the classes differ \emph{only} in the right arm: in the \emph{motion} task one class replays the arm at $25\%$ speed (its shape is real), in the \emph{shape} task one class thickens the arm $3\times$ about its bones (its motion is untouched). We keep the videos in which the right arm carries at least 30\% of the skeleton's motion (\NumXTrain\ training and \NumXTest\ test videos, one centred clip each) and train one P4Transformer per task; its accuracy over both classes of the test clips is \NumXMotAcc\ / \NumXGeoAcc\,\% (motion / shape task). In each task we explain the class whose defining property is present and removable by the interventions---normal-speed motion (removed by \frz) and the thick shape (removed by the proxy). We explain the first \NumXEval\ test clips and pool the positive $\Smot$ and $\Sgeo$ over their arm points. For the \emph{same} arm, the motion share of \method's importance is \NumXMotShare\ for the motion model (95\% bootstrap CI [\NumXMotLo, \NumXMotHi]; context estimator \NumXMotShareCtx) and \NumXGeoShare\ for the shape model ([\NumXGeoLo, \NumXGeoHi]; context \NumXGeoShareCtx)---a gap of [\NumXGapLo, \NumXGapHi]---whereas the model-blind motion prior marks the arm identically in both tasks (EBPG \NumXMotPriorEBPG\ vs.\ \NumXGeoPriorEBPG). Across \NumXSeedsTotal\ training seeds of the two task models, the motion-task model leaves chance in only \NumXSeedsTrained; in all \NumXSeedsSameDir\ of these the arm's motion share is higher in the motion task, with an interval that excludes zero in \NumXSeedsCIExcl\ (Tab.~\ref{tab:seeds}). The split is not all-or-nothing on real models---the shape model still reacts to freezing the arm, whose frozen pose changes its per-frame configuration---but it moves with the cue, which no single map can show. Repeating the test with the convolutional PSTNet (\NumXPstMotAcc\ / \NumXPstGeoAcc\,\% accuracy) gives a cleaner split, \NumXPstMotShare\ vs.\ \NumXPstGeoShare\ (gap [\NumXPstGapLo, \NumXPstGapHi]), and the state-space Mamba4D (\NumXMbaMotAcc\ / \NumXMbaGeoAcc\,\%) one similar to P4Transformer, \NumXMbaMotShare\ vs.\ \NumXMbaGeoShare\ (gap [\NumXMbaGapLo, \NumXMbaGapHi]). On these two backbones the shape model still draws more than half of the arm's importance from motion, in line with their motion-heavy maps on MSR (whole-body motion share \NumMBAKcaCtxMotShareAll\ for Mamba4D, \NumPFTKcaCtxMotShareAll\ for P4Transformer): the audit shows which cue a model responds to, not that it uses that cue alone. Fig.~\ref{fig:dissoc} summarises all three next to the fully controlled setting of Appendix~\ref{supp:controlled} (\NumCtrlMot\ vs.\ \NumCtrlShape).

\begin{figure}[t]
\centering
\includegraphics[width=0.9\linewidth]{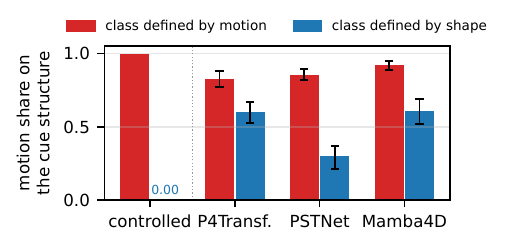}
\caption{\textbf{Double dissociation:} motion share on the cue structure when the class is defined by its motion (red) or by its shape (blue), local estimator. Controlled task: pooled over 3 seeds $\times$ 20 clips. Real crossover: \NumXEval\ test clips per task and backbone, with 95\% bootstrap intervals.}
\label{fig:dissoc}
\end{figure}

\subsection{Ablation, sanity and cost (Q4)}
\label{sec:exp-ablation}
\paragraph{Components (Tab.~\ref{tab:ablation} in the appendix).} \emph{Redundancy:} with a fine partition ($m{=}24$) a body part spans several regions and the local necessity collapses (deletion AUC \NumAblLocalFineDel), while the context estimator keeps it (\NumAblCtxFineDel)---the failure mode it was designed for. \emph{The $(0,0)$ corner:} deleting the region instead of using the static proxy aligns the necessity with the deletion metric (local \NumAblLocalDel$\to$\NumAblLocalBgDel; context \NumAblCtxDel$\to$\NumAblCtxBgDel, the best of all variants) but weakens the motion term's localisation (EBPG \NumAblLocalEBPG$\to$\NumAblLocalBgEBPG); in the controlled study, where the answer is known, the same change makes a pure-shape model look \NumCtrlBgShape\% motion-driven (Appendix~\ref{supp:controlled}). \emph{The order:} the motion-first scores absorb the interaction into $\Smot$ (motion share \NumAblLocalMotionFirstShare\ vs.\ \NumAblLocalShare\ symmetric). $S{=}64$ samples already give \NumAblCtxSsixtyfourDel\ (vs.\ \NumAblCtxDel\ at $S{=}256$). Anchoring \frz\ and the proxy at the middle instead of the first frame barely changes the split (motion share \NumAblLocalMidShare\ vs.\ \NumAblLocalShare) and does not hurt faithfulness (Tab.~\ref{tab:ablation}). The interaction is \emph{not} negligible on real models ($|\Delta_{\mathrm{int}}|/|\Nec|$ in the table, against \NumCtrlProxyDint\ for the controlled network), which is exactly why we report order-symmetric scores and the residual rather than assume separability.
\paragraph{Sanity.} Under the cascading randomisation of \citet{adebayo2018sanity}, applied to P4Transformer from the head down on 20 clips (Fig.~\ref{fig:sanity} in the appendix), the rank correlation of \method's map with the trained model's drops to \NumSanHeadKca\ as soon as the head is re-initialised and stays within $\pm0.2$ (\NumSanKca\ with every block randomised): the explanation depends on the learned weights. Grad$\times$Input keeps a correlation of \NumSanGxI\ even then---much of it reflects the input rather than the model. On the region with the largest $\Smot$ (\NumCrossClips\ clips), freezing it changes the prediction in \NumDspFreezeFlip\% of the clips and reversing its trajectory (\dsp) in \NumDspReverseFlip\%: most motion evidence is about \emph{whether} a part moves, less about \emph{which way}.

\paragraph{Stability under re-sampling.} A sensor never returns the same points twice, so we re-sample each of the \NumCrossClips\ clips \NumStabSamplings\ times, explain every sampling, map each explanation onto the points of the first one and rank-correlate the two (Tab.~\ref{tab:stability} in the appendix). The prediction stays the same in \NumStabPredSame\% of the samplings. \method's maps are the less stable ones point by point: \NumStabKcaCtxPoint\ for the context estimator against \NumStabIGPoint\ for Integrated Gradients and \NumStabOcclPoint\ for region occlusion, because the regions are re-fitted to every sampling. At the level of body parts (each map averaged over the 19 skeleton bones) the gap narrows (\NumStabKcaCtxPart\ vs.\ \NumStabIGPart), and the whole-body motion share varies little between samplings (standard deviation \NumStabKcaCtxShareSD). Averaging the maps of two samplings, which costs only forward passes, raises the context estimator to \NumStabKcaCtxAvgPoint\ (points) and \NumStabKcaCtxAvgPart\ (parts).

\paragraph{Cost.} On one A100 GPU (mean over the \NumPFTClips\ clips), the local estimator takes \NumPFTKcaTime\,s per clip and the context estimator \NumPFTKcaCtxTime\,s ($S{=}256$ random states, batched), comparable to OccAM (\NumPFTOccamTime\,s) and Integrated Gradients (\NumPFTIGTime\,s). All passes are independent forward queries, so the cost falls with batch throughput and needs no backward pass.

\subsection{Qualitative results}
\label{sec:exp-qual}
Fig.~\ref{fig:teaser} and Fig.~\ref{fig:qual} in the appendix overlay $\Smot$ and $\Sgeo$ on real test clips. On \emph{draw circle} all three backbones put their strongest motion importance on the drawing hand, and on \emph{forward kick} P4Transformer and PSTNet put it on the kicking leg. \emph{High arm wave} is different: PSTNet marks the raised hand, but P4Transformer and Mamba4D put their motion importance on the torso rather than on the waving arm. Geometry importance spreads more widely over the body on every backbone. Per-action statistics are in Tab.~\ref{tab:per-class}.

\section{Conclusion}
\label{sec:conclusion}

\method explains 4D point-cloud video models with forward queries alone and needs no point correspondences. It asks a question that no prior 4D explanation can pose: does a decision depend on \emph{how} a structure moves or on \emph{what shape} it has? Three findings support it.
\textbf{(1) The split is correct.} Where the answer is known by construction, the motion share on the cue structure is \NumCtrlMot\ for a motion-defined and \NumCtrlShape\ for a shape-defined class (Appendix~\ref{supp:controlled}). On real clips, when two models see the same arm but are trained on its motion or on its shape, the motion share of the arm's importance follows the cue on all three backbones (\NumXMotShare\ vs.\ \NumXGeoShare\ on P4Transformer, \NumXPstMotShare\ vs.\ \NumXPstGeoShare\ on PSTNet, \NumXMbaMotShare\ vs.\ \NumXMbaGeoShare\ on Mamba4D; every gap's 95\% interval excludes zero), whereas a model-blind motion map marks the arm identically in both tasks and a single attribution map has no split at all.
\textbf{(2) The split costs no faithfulness.} The necessity map has the lowest deletion AUC on P4Transformer (significantly so on the full test set) and on PSTNet, ties with the best baseline on Mamba4D, and loses its correlation with the trained model's map once the weights are randomised (\NumSanKca, against \NumSanGxI\ for Grad$\times$Input).
\textbf{(3) It is deployable.} The same code explains convolutional, attention-based and attention-free backbones, while attention read-outs exist for only one of them, at \NumPFTKcaTime--\NumPFTKcaCtxTime\,s per clip, comparable to Integrated Gradients.
\method is not the best map on every axis: white-box Integrated Gradients localises the moving body parts better, and its point-level maps are more stable under sensor re-sampling. What \method adds is the motion/geometry split, which no gradient, attention or occlusion map provides.

\paragraph{Practical guidance.}
For auditing a 4D model before deployment we recommend the following. (i) Use the context-averaged estimator for the final report and the local one for quick inspection: the local one needs $3m{+}1$ passes (\NumPFTKcaTime\,s per clip on one A100 GPU) and the context one $S$ passes (\NumPFTKcaCtxTime\,s). (ii) Report $\Delta_{\mathrm{int}}$ alongside $\Smot$ and $\Sgeo$. On real models it is large, so a single-order split would be arbitrary. (iii) Average the maps of a few re-samplings of the same clip when the explanation must be stable; it costs only forward passes. (iv) When the intended cue is known, test it directly with a crossover-style pair of tasks. A model whose arm importance stays motion-dominated on a shape-defined task, as P4Transformer's and Mamba4D's do here, also responds to a cue that does not define its classes.

\paragraph{Limitations.}
(i) No pretrained weights exist, so we trained the backbones from scratch. P4Transformer comes close to its published accuracy, while PSTNet and Mamba4D fall short of theirs. A laptop-scale training gives the same picture (Appendix~\ref{supp:scale}). At full scale the P4Transformer motion-task model trains unreliably: it leaves chance in only \NumXSeedsTrained\ of \NumXSeedsTotal\ training seeds (Tab.~\ref{tab:seeds}). Further datasets (NTU RGB+D, HOI4D) remain to be tested. (ii) Skeleton-derived ground truth is approximate. It marks where motion \emph{is}, not where the model \emph{should} look, which is why our primary evidence for the decomposition is the crossover rather than localisation. (iii) The decomposition's semantics require the proxy corner. With deletion at that corner the necessity map aligns better with removal-based metrics, but presence leaks into the motion term (Tab.~\ref{tab:ablation}). (iv) Regions come from a k-means tracker that can drift under fast or occluded motion, and the scores are only as meaningful as the regions they are computed on; because the regions are re-fitted to every sampling, \method's point-level maps are less stable under sensor re-sampling than gradient maps.

\paragraph{Future work.}
Attributing 4D occupancy \emph{forecasts}~\cite{tian2023occ3d,ma2024cam4docc}, so far served only by uncertainty estimates~\cite{reliocc2025,kalble2025evocc,occuq2025}, to the past motion that caused them, and interventional probes of adversarial behaviour on 4D models~\cite{tan2023onepoint,li2021foolinglidar}.

\label{maintext-end}

\FloatBarrier
\bibliography{main}

\begin{thebibliography}{65}
\providecommand{\natexlab}[1]{#1}
\providecommand{\url}[1]{\texttt{#1}}
\expandafter\ifx\csname urlstyle\endcsname\relax
  \providecommand{\doi}[1]{doi: #1}\else
  \providecommand{\doi}{doi: \begingroup \urlstyle{rm}\Url}\fi

\bibitem[Abnar and Zuidema(2020)]{abnar2020rollout}
Samira Abnar and Willem Zuidema.
\newblock Quantifying attention flow in transformers.
\newblock In \emph{ACL}, 2020.

\bibitem[Adebayo et~al.(2018)Adebayo, Gilmer, Muelly, Goodfellow, Hardt, and Kim]{adebayo2018sanity}
Julius Adebayo, Justin Gilmer, Michael Muelly, Ian Goodfellow, Moritz Hardt, and Been Kim.
\newblock Sanity checks for saliency maps.
\newblock In \emph{NeurIPS}, 2018.

\bibitem[Ancona et~al.(2018)Ancona, Ceolini, {\"O}ztireli, and Gross]{ancona2018sensitivity}
Marco Ancona, Enea Ceolini, Cengiz {\"O}ztireli, and Markus Gross.
\newblock Towards better understanding of gradient-based attribution methods for deep neural networks.
\newblock In \emph{ICLR}, 2018.

\bibitem[Banzhaf~III(1965)]{banzhaf1965}
John~F Banzhaf~III.
\newblock Weighted voting doesn't work: A mathematical analysis.
\newblock \emph{Rutgers Law Review}, 19:\penalty0 317--343, 1965.

\bibitem[Ben-Shabat et~al.(2024)Ben-Shabat, Shrout, and Gould]{benshabat2024_3dinaction}
Yizhak Ben-Shabat, Oren Shrout, and Stephen Gould.
\newblock {3DInAction}: Understanding human actions in 3d point clouds.
\newblock In \emph{CVPR}, pages 19978--19987, 2024.

\bibitem[Chakraborty et~al.(2026)Chakraborty, Karagoz, and Meratnia]{chakraborty2026benchmark}
Asmita Chakraborty, Gizem Karagoz, and Nirvana Meratnia.
\newblock A cross-domain benchmark of intrinsic and post hoc explainability for {3D} deep learning models.
\newblock \emph{Journal of Imaging}, 12\penalty0 (2):\penalty0 63, 2026.

\bibitem[Chefer et~al.(2021)Chefer, Gur, and Wolf]{chefer2021transformer}
Hila Chefer, Shir Gur, and Lior Wolf.
\newblock Transformer interpretability beyond attention visualization.
\newblock In \emph{CVPR}, 2021.

\bibitem[De~Vries et~al.(2025)De~Vries, Naidoo, Fourkioti, Dent, Curry, Dunsby, and Bakal]{devries2025pointmil}
Matt De~Vries, Reed Naidoo, Olga Fourkioti, Lucas~G Dent, Nathan Curry, Chris Dunsby, and Chris Bakal.
\newblock Interpretable point cloud classification using multiple instance learning.
\newblock In \emph{ICCV}, pages 22209--22220, 2025.

\bibitem[DeYoung et~al.(2020)DeYoung, Jain, Rajani, Lehman, Xiong, Socher, and Wallace]{deyoung2020eraser}
Jay DeYoung, Sarthak Jain, Nazneen~Fatema Rajani, Eric Lehman, Caiming Xiong, Richard Socher, and Byron~C Wallace.
\newblock {ERASER}: A benchmark to evaluate rationalized {NLP} models.
\newblock In \emph{ACL}, 2020.

\bibitem[Dunn(1961)]{dunn1961}
Olive~Jean Dunn.
\newblock Multiple comparisons among means.
\newblock \emph{Journal of the American Statistical Association}, 56\penalty0 (293):\penalty0 52--64, 1961.

\bibitem[Efron(1979)]{efron1979bootstrap}
Bradley Efron.
\newblock Bootstrap methods: Another look at the jackknife.
\newblock \emph{The Annals of Statistics}, 7\penalty0 (1):\penalty0 1--26, 1979.

\bibitem[Fan et~al.(2021{\natexlab{a}})Fan, Yang, and Kankanhalli]{fan2021p4transformer}
Hehe Fan, Yi~Yang, and Mohan Kankanhalli.
\newblock Point 4d transformer networks for spatio-temporal modeling in point cloud videos.
\newblock In \emph{CVPR}, 2021{\natexlab{a}}.

\bibitem[Fan et~al.(2021{\natexlab{b}})Fan, Yang, and Kankanhalli]{fan2021pstnet}
Hehe Fan, Yi~Yang, and Mohan Kankanhalli.
\newblock {PSTNet}: Point spatio-temporal convolution on point cloud sequences.
\newblock In \emph{ICLR}, 2021{\natexlab{b}}.

\bibitem[Fan et~al.(2023)Fan, Yang, and Kankanhalli]{fan2023psttransformer}
Hehe Fan, Yi~Yang, and Mohan Kankanhalli.
\newblock Point spatio-temporal transformer networks for point cloud video modeling.
\newblock \emph{IEEE TPAMI}, 2023.

\bibitem[Fiebelman et~al.(2025)Fiebelman, Cohen, Morgenstern, Hedman, and Averbuch-Elor]{fiebelman2025_4legs}
Gal Fiebelman, Tamir Cohen, Ayellet Morgenstern, Peter Hedman, and Hadar Averbuch-Elor.
\newblock {4-LEGS}: 4d language embedded gaussian splatting.
\newblock \emph{Computer Graphics Forum}, 44\penalty0 (2):\penalty0 e70085, 2025.

\bibitem[Fong et~al.(2019)Fong, Patrick, and Vedaldi]{fong2019extremal}
Ruth Fong, Mandela Patrick, and Andrea Vedaldi.
\newblock Understanding deep networks via extremal perturbations and smooth masks.
\newblock In \emph{ICCV}, 2019.

\bibitem[Galus et~al.(2026)Galus, Farganus, Zapala, Czachorowski, Borycki, Spurek, and Syga]{galus2026xsplain}
Dominik Galus, Julia Farganus, Tymoteusz Zapala, Miko{\l}aj Czachorowski, Piotr Borycki, Przemys{\l}aw Spurek, and Piotr Syga.
\newblock {XSPLAIN}: {XAI}-enabling splat-based prototype learning for attribute-aware interpretability.
\newblock \emph{arXiv preprint arXiv:2602.10239}, 2026.

\bibitem[Goyal et~al.(2019)Goyal, Feder, Shalit, and Kim]{goyal2019cace}
Yash Goyal, Amir Feder, Uri Shalit, and Been Kim.
\newblock Explaining classifiers with causal concept effect ({CaCE}).
\newblock \emph{arXiv preprint arXiv:1907.07165}, 2019.

\bibitem[Grabisch and Roubens(1999)]{grabisch1999interaction}
Michel Grabisch and Marc Roubens.
\newblock An axiomatic approach to the concept of interaction among players in cooperative games.
\newblock \emph{International Journal of Game Theory}, 28\penalty0 (4):\penalty0 547--565, 1999.

\bibitem[Gu and Dao(2024)]{gu2024mamba}
Albert Gu and Tri Dao.
\newblock Mamba: Linear-time sequence modeling with selective state spaces.
\newblock In \emph{COLM}, 2024.

\bibitem[Heidrich et~al.(2025)Heidrich, Beemelmanns, Nekrasov, Leibe, and Eckstein]{occuq2025}
Severin Heidrich, Till Beemelmanns, Alexey Nekrasov, Bastian Leibe, and Lutz Eckstein.
\newblock {OCCUQ}: Exploring efficient uncertainty quantification for 3d occupancy prediction.
\newblock In \emph{ICRA}, 2025.
\newblock arXiv:2503.10605.

\bibitem[Jain and Wallace(2019)]{jain2019attention}
Sarthak Jain and Byron~C Wallace.
\newblock Attention is not explanation.
\newblock In \emph{NAACL}, 2019.

\bibitem[Janzing et~al.(2020)Janzing, Minorics, and Bl{\"o}baum]{janzing2020feature}
Dominik Janzing, Lenon Minorics, and Patrick Bl{\"o}baum.
\newblock Feature relevance quantification in explainable {AI}: A causal problem.
\newblock In \emph{AISTATS}, 2020.

\bibitem[K{\"a}lble et~al.(2025)K{\"a}lble, Wirges, Tatarchenko, and Ilg]{kalble2025evocc}
Jonas K{\"a}lble, Sascha Wirges, Maxim Tatarchenko, and Eddy Ilg.
\newblock {EvOcc}: Accurate semantic occupancy for automated driving using evidence theory.
\newblock In \emph{CVPR}, 2025.

\bibitem[Knab et~al.(2025)Knab, Marton, Schubert, Guggiana, and Bartelt]{knab2025motif}
Patrick Knab, Sascha Marton, Philipp~J Schubert, Drago Guggiana, and Christian Bartelt.
\newblock Concepts in motion: Temporal concept bottleneck model for interpretable video classification.
\newblock \emph{arXiv preprint arXiv:2509.20899}, 2025.

\bibitem[Kwon et~al.(2025)Kwon, Cho, and Kim]{kwon2025instruct4dgs}
Joohyun Kwon, Hanbyel Cho, and Junmo Kim.
\newblock Efficient dynamic scene editing via 4d gaussian-based static-dynamic separation.
\newblock In \emph{CVPR}, 2025.
\newblock arXiv:2502.02091.

\bibitem[Lee et~al.(2025)Lee, Lee, Park, Kim, and Choi]{lee2025dance}
Jongseo Lee, Wooil Lee, Gyeong-Moon Park, Seong~Tae Kim, and Jinwoo Choi.
\newblock Disentangled concepts speak louder than words: Explainable video action recognition.
\newblock In \emph{NeurIPS}, 2025.

\bibitem[Li et~al.(2023)Li, Liu, Zhang, and Li]{li2023staticbias}
Haoxin Li, Yuan Liu, Hanwang Zhang, and Boyang Li.
\newblock Mitigating and evaluating static bias of action representations in the background and the foreground.
\newblock In \emph{ICCV}, pages 19911--19923, 2023.

\bibitem[Li et~al.(2025)Li, Zhou, Zhou, Song, Herter, Qin, Huang, and Pfister]{li2025_4dlangsplat}
Wanhua Li, Renping Zhou, Jiawei Zhou, Yingwei Song, Johannes Herter, Minghan Qin, Gao Huang, and Hanspeter Pfister.
\newblock {4D LangSplat}: 4d language gaussian splatting via multimodal large language models.
\newblock In \emph{CVPR}, 2025.
\newblock arXiv:2503.10437.

\bibitem[Li et~al.(2010)Li, Zhang, and Liu]{li2010msraction3d}
Wanqing Li, Zhengyou Zhang, and Zicheng Liu.
\newblock Action recognition based on a bag of 3d points.
\newblock In \emph{CVPR Workshops}, 2010.

\bibitem[Li et~al.(2021)Li, Wen, Juefei-Xu, and Feng]{li2021foolinglidar}
Yiming Li, Congcong Wen, Felix Juefei-Xu, and Chen Feng.
\newblock Fooling {LiDAR} perception via adversarial trajectory perturbation.
\newblock In \emph{ICCV}, 2021.

\bibitem[Liu et~al.(2025)Liu, Han, Liu, Aviles-Rivero, Jiang, Liu, and Wang]{liu2025mamba4d}
Jiuming Liu, Jinru Han, Lihao Liu, Angelica~I Aviles-Rivero, Chaokang Jiang, Zhe Liu, and Hesheng Wang.
\newblock {Mamba4D}: Efficient 4d point cloud video understanding with disentangled spatial-temporal state space models.
\newblock In \emph{CVPR}, 2025.
\newblock arXiv:2405.14338.

\bibitem[Liu et~al.(2019)Liu, Yan, and Bohg]{liu2019meteornet}
Xingyu Liu, Mengyuan Yan, and Jeannette Bohg.
\newblock {MeteorNet}: Deep learning on dynamic 3d point cloud sequences.
\newblock In \emph{ICCV}, 2019.

\bibitem[Lundberg and Lee(2017)]{lundberg2017shap}
Scott~M Lundberg and Su-In Lee.
\newblock A unified approach to interpreting model predictions.
\newblock In \emph{NeurIPS}, 2017.

\bibitem[Ma et~al.(2024)Ma, Chen, Huang, Xu, Luo, Xu, Gu, Ai, and Wang]{ma2024cam4docc}
Junyi Ma, Xieyuanli Chen, Jiawei Huang, Jingyi Xu, Zhen Luo, Jintao Xu, Weihao Gu, Rui Ai, and Hesheng Wang.
\newblock {Cam4DOcc}: Benchmark for camera-only 4d occupancy forecasting in autonomous driving applications.
\newblock In \emph{CVPR}, pages 21486--21495, 2024.

\bibitem[Paszke et~al.(2019)Paszke, Gross, Massa, Lerer, Bradbury, Chanan, Killeen, Lin, Gimelshein, Antiga, Desmaison, K{\"o}pf, Yang, DeVito, Raison, Tejani, Chilamkurthy, Steiner, Fang, Bai, and Chintala]{paszke2019pytorch}
Adam Paszke, Sam Gross, Francisco Massa, Adam Lerer, James Bradbury, Gregory Chanan, Trevor Killeen, Zeming Lin, Natalia Gimelshein, Luca Antiga, Alban Desmaison, Andreas K{\"o}pf, Edward Yang, Zachary DeVito, Martin Raison, Alykhan Tejani, Sasank Chilamkurthy, Benoit Steiner, Lu~Fang, Junjie Bai, and Soumith Chintala.
\newblock {PyTorch}: An imperative style, high-performance deep learning library.
\newblock In \emph{NeurIPS}, 2019.

\bibitem[Pearl(2009)]{pearl2009causality}
Judea Pearl.
\newblock \emph{Causality: Models, Reasoning, and Inference}.
\newblock Cambridge University Press, 2 edition, 2009.

\bibitem[Petsiuk et~al.(2018)Petsiuk, Das, and Saenko]{petsiuk2018rise}
Vitali Petsiuk, Abir Das, and Kate Saenko.
\newblock {RISE}: Randomized input sampling for explanation of black-box models.
\newblock In \emph{BMVC}, 2018.

\bibitem[Price and Damen(2020)]{price2020playfair}
Will Price and Dima Damen.
\newblock Play fair: Frame contributions in video models.
\newblock In \emph{ACCV}, 2020.

\bibitem[Qi et~al.(2017{\natexlab{a}})Qi, Su, Mo, and Guibas]{qi2017pointnet}
Charles~R Qi, Hao Su, Kaichun Mo, and Leonidas~J Guibas.
\newblock {PointNet}: Deep learning on point sets for 3d classification and segmentation.
\newblock In \emph{CVPR}, 2017{\natexlab{a}}.

\bibitem[Qi et~al.(2017{\natexlab{b}})Qi, Yi, Su, and Guibas]{qi2017pointnetpp}
Charles~R Qi, Li~Yi, Hao Su, and Leonidas~J Guibas.
\newblock {PointNet++}: Deep hierarchical feature learning on point sets in a metric space.
\newblock In \emph{NeurIPS}, 2017{\natexlab{b}}.

\bibitem[Ribeiro et~al.(2016)Ribeiro, Singh, and Guestrin]{ribeiro2016lime}
Marco~Tulio Ribeiro, Sameer Singh, and Carlos Guestrin.
\newblock ``why should i trust you?'': Explaining the predictions of any classifier.
\newblock In \emph{KDD}, 2016.

\bibitem[Saha et~al.(2024)Saha, Gupta, Ankireddy, Chahine, and Ghosh]{videotcav2024}
Avinab Saha, Shashank Gupta, Sravan~Kumar Ankireddy, Karl Chahine, and Joydeep Ghosh.
\newblock Exploring explainability in video action recognition.
\newblock In \emph{CVPR Workshops}, pages 8176--8181, 2024.

\bibitem[Schinagl et~al.(2022)Schinagl, Krispel, Possegger, Roth, and Bischof]{schinagl2022occam}
David Schinagl, Georg Krispel, Horst Possegger, Peter~M Roth, and Horst Bischof.
\newblock {OccAM's Laser}: Occlusion-based attribution maps for 3d object detectors on {LiDAR} data.
\newblock In \emph{CVPR}, 2022.

\bibitem[Selvaraju et~al.(2017)Selvaraju, Cogswell, Das, Vedantam, Parikh, and Batra]{selvaraju2017gradcam}
Ramprasaath~R Selvaraju, Michael Cogswell, Abhishek Das, Ramakrishna Vedantam, Devi Parikh, and Dhruv Batra.
\newblock {Grad-CAM}: Visual explanations from deep networks via gradient-based localization.
\newblock In \emph{ICCV}, 2017.

\bibitem[Smilkov et~al.(2017)Smilkov, Thorat, Kim, Vi{\'e}gas, and Wattenberg]{smilkov2017smoothgrad}
Daniel Smilkov, Nikhil Thorat, Been Kim, Fernanda Vi{\'e}gas, and Martin Wattenberg.
\newblock {SmoothGrad}: removing noise by adding noise.
\newblock \emph{arXiv preprint arXiv:1706.03825}, 2017.

\bibitem[Sundararajan et~al.(2017)Sundararajan, Taly, and Yan]{sundararajan2017ig}
Mukund Sundararajan, Ankur Taly, and Qiqi Yan.
\newblock Axiomatic attribution for deep networks.
\newblock In \emph{ICML}, 2017.

\bibitem[Tan and Kotthaus(2023)]{tan2023onepoint}
Hanxiao Tan and Helena Kotthaus.
\newblock Explainability-aware one point attack for point cloud neural networks.
\newblock In \emph{WACV}, 2023.

\bibitem[Tayyub et~al.(2022)Tayyub, Sarmad, and Sch{\"o}nborn]{tayyub2022ape}
Jawad Tayyub, Muhammad Sarmad, and Nicolas Sch{\"o}nborn.
\newblock Explaining deep neural networks for point clouds using gradient-based visualizations.
\newblock In \emph{ACCV}, 2022.

\bibitem[Tian et~al.(2023)Tian, Jiang, Yun, Mao, Yang, Wang, Wang, and Zhao]{tian2023occ3d}
Xiaoyu Tian, Tao Jiang, Longfei Yun, Yucheng Mao, Huitong Yang, Yue Wang, Yilun Wang, and Hang Zhao.
\newblock {Occ3D}: A large-scale 3d occupancy prediction benchmark for autonomous driving.
\newblock In \emph{NeurIPS}, 2023.

\bibitem[Varshney et~al.(2025)Varshney, Lucieri, Balada, Ahmed, and Dengel]{ldvice2025}
Payal Varshney, Adriano Lucieri, Christoph Balada, Sheraz Ahmed, and Andreas Dengel.
\newblock {LD-ViCE}: Latent diffusion model for video counterfactual explanations.
\newblock \emph{arXiv preprint arXiv:2509.08422}, 2025.

\bibitem[Virtanen et~al.(2020)Virtanen, Gommers, Oliphant, et~al.]{virtanen2020scipy}
Pauli Virtanen, Ralf Gommers, Travis~E. Oliphant, et~al.
\newblock {SciPy} 1.0: Fundamental algorithms for scientific computing in {Python}.
\newblock \emph{Nature Methods}, 17\penalty0 (3):\penalty0 261--272, 2020.

\bibitem[Wang et~al.(2020)Wang, Wang, Du, Yang, Zhang, Ding, Mardziel, and Hu]{wang2020scorecam}
Haofan Wang, Zifan Wang, Mengnan Du, Fan Yang, Zijian Zhang, Sirui Ding, Piotr Mardziel, and Xia Hu.
\newblock {Score-CAM}: Score-weighted visual explanations for convolutional neural networks.
\newblock In \emph{CVPR Workshops}, 2020.

\bibitem[Wang et~al.(2025)Wang, Wang, Yu, Li, Feng, Chen, and Zhu]{reliocc2025}
Song Wang, Zhongdao Wang, Jiawei Yu, Wentong Li, Bailan Feng, Junbo Chen, and Jianke Zhu.
\newblock Reliable and calibrated semantic occupancy prediction by hybrid uncertainty learning.
\newblock In \emph{IJCAI}, pages 1973--1981, 2025.

\bibitem[Wang et~al.(2026)Wang, Sun, Wang, Zhang, Li, Yin, and Li]{wang2026mosaic}
Wei Wang, Yiding Sun, Yuyan Wang, Zhuoyue Zhang, Zhengqiao Li, Dongfu Yin, and Chen Li.
\newblock Motion-saliency complementary masked modeling for point cloud video understanding.
\newblock \emph{arXiv preprint arXiv:2608.30279}, 2026.

\bibitem[Wang et~al.(2019)Wang, Sun, Liu, Sarma, Bronstein, and Solomon]{wang2019dgcnn}
Yue Wang, Yongbin Sun, Ziwei Liu, Sanjay~E Sarma, Michael~M Bronstein, and Justin~M Solomon.
\newblock Dynamic graph {CNN} for learning on point clouds.
\newblock \emph{ACM Transactions on Graphics (TOG)}, 2019.

\bibitem[Wang and Liu(2025)]{wang2025staa}
Zerui Wang and Yan Liu.
\newblock {STAA}: Spatio-temporal attention attribution for real-time interpreting transformer-based {AI} video models.
\newblock \emph{IEEE Access}, 13:\penalty0 101647--101661, 2025.

\bibitem[Watson et~al.(2021)Watson, Gultchin, Taly, and Floridi]{watson2021lens}
David~S Watson, Limor Gultchin, Ankur Taly, and Luciano Floridi.
\newblock Local explanations via necessity and sufficiency: Unifying theory and practice.
\newblock In \emph{UAI}, 2021.

\bibitem[Wen et~al.(2022)Wen, Liu, Huang, Duan, and Yi]{wen2022pptr}
Hao Wen, Yunze Liu, Jingwei Huang, Bo~Duan, and Li~Yi.
\newblock Point primitive transformer for long-term 4d point cloud video understanding.
\newblock In \emph{ECCV}, 2022.

\bibitem[Wilcoxon(1945)]{wilcoxon1945}
Frank Wilcoxon.
\newblock Individual comparisons by ranking methods.
\newblock \emph{Biometrics Bulletin}, 1\penalty0 (6):\penalty0 80--83, 1945.

\bibitem[Wu et~al.(2024)Wu, Jiang, Wang, Liu, Liu, Qiao, Ouyang, He, and Zhao]{wu2024ptv3}
Xiaoyang Wu, Li~Jiang, Peng-Shuai Wang, Zhijian Liu, Xihui Liu, Yu~Qiao, Wanli Ouyang, Tong He, and Hengshuang Zhao.
\newblock Point transformer {V3}: Simpler, faster, stronger.
\newblock In \emph{CVPR}, 2024.

\bibitem[Wu et~al.(2026)Wu, Paschalidou, Gao, Torralba, Leal-Taix{\'e}, Russakovsky, Fidler, and Lorraine]{wu2026motive}
Xindi Wu, Despoina Paschalidou, Jun Gao, Antonio Torralba, Laura Leal-Taix{\'e}, Olga Russakovsky, Sanja Fidler, and Jonathan Lorraine.
\newblock Motion attribution for video generation.
\newblock In \emph{ICML}, 2026.

\bibitem[Zhang et~al.(2018)Zhang, Bargal, Lin, Brandt, Shen, and Sclaroff]{zhang2018pointing}
Jianming Zhang, Sarah~Adel Bargal, Zhe Lin, Jonathan Brandt, Xiaohui Shen, and Stan Sclaroff.
\newblock Top-down neural attention by excitation backprop.
\newblock \emph{International Journal of Computer Vision (IJCV)}, 2018.

\bibitem[Zhang et~al.(2024)Zhang, Song, Lee, Yang, Peng, Chellappa, and Fan]{zhang2024lp3dgs}
Zhaoliang Zhang, Tianchen Song, Yongjae Lee, Li~Yang, Cheng Peng, Rama Chellappa, and Deliang Fan.
\newblock {LP-3DGS}: Learning to prune 3d gaussian splatting.
\newblock In \emph{NeurIPS}, 2024.
\newblock arXiv:2405.18784.

\bibitem[Zheng et~al.(2019)Zheng, Chen, Yuan, Li, and Ren]{zheng2019pointcloudsaliency}
Tianhang Zheng, Changyou Chen, Junsong Yuan, Bo~Li, and Kui Ren.
\newblock {PointCloud} saliency maps.
\newblock In \emph{ICCV}, 2019.

\end{thebibliography}

\appendix

\section{Notation and pseudocode}
\label{supp:notation}
Tab.~\ref{tab:notation} lists the notation, Tab.~\ref{tab:assump} the assumptions behind each result, and Alg.~\ref{alg:kca} the procedure.

\begin{table*}[h]
\centering\small
\begin{tabular}{ll}
\toprule
$X=\{P_t\}_{t=1}^{T}$ & point-cloud video; $P_t\!\subset\!\real^3$, sampled per frame \\
$f,\ \varphi(X){=}f(X)_c$ & model; target logit of class $c$ \\
$\setR=\{\setR_t\}_t$ & a tracked region: its point set in each frame \\
$c_t(\setR)$ & centroid of $\setR_t$ \\
$\doop{\frz,\setR}$ & freeze: $\setR$ static at its $t_0$ shape ($M{=}0$, $G{=}1$) \\
$\doop{\ablgeo,\setR}$ & shape-free proxy moving along $c_t$ ($M{=}1$, $G{=}0$) \\
$\doop{\abl,\setR}$ & shape-free proxy static at $c_{t_0}$ ($M{=}0$, $G{=}0$) \\
$\Nec,\ \Smot,\Sgeo,\ \Delta_{\mathrm{int}}$ & necessity; motion, geometry importance; interaction \\
\bottomrule
\end{tabular}
\caption{Notation.}
\label{tab:notation}
\end{table*}

\begin{table*}[h]
\centering\small
\begin{tabular}{lll}
\toprule
& Assumption & Needed by \\
\midrule
\textbf{I} & interventional validity (operators rewrite only $\setR_t$) & Props.~\ref{prop:ace},~\ref{prop:pnps},~\ref{prop:context} \\
\textbf{S} & separability \emph{(measured by $\Delta_{\mathrm{int}}$, not assumed)} & Prop.~\ref{prop:ace} (marginal ACE only) \\
\textbf{M} & monotonicity \& exogeneity & Prop.~\ref{prop:pnps} (identification) \\
\bottomrule
\end{tabular}
\caption{Assumptions and the results that require them.}
\label{tab:assump}
\end{table*}

\begin{algorithm}[h]
\caption{\method: Kinematic Causal Attribution}
\label{alg:kca}
\begin{algorithmic}[1]
\Require clip $X$, model $f$, class $c$, tracked regions $\{\setR^j\}_{j=1}^{m}$
\Ensure motion field $\Smot$, geometry field $\Sgeo$, interaction $\Delta_{\mathrm{int}}$
\State precompute for all regions: frozen sets $\setR^j_{t_0}$, proxy paths $c_t(\setR^j)$
\If{local}
  \State evaluate $\varphi$ at $X$ and 3 corners per region \Comment{$3m{+}1$ passes}
  \State Eq.~\eqref{eq:sym}, Eq.~\eqref{eq:dint} per region
\Else
  \State evaluate $\varphi$ on $S$ random joint states \Comment{$S$ passes}
  \State solve Eq.~\eqref{eq:context}; $\Smot{=}\mu{+}\iota/2$, $\Sgeo{=}\gamma{+}\iota/2$
\EndIf
\State \Return per-point fields: each point receives its region's scores
\end{algorithmic}
\end{algorithm}

\section{Full proofs}
\label{supp:proofs}
Notation: $\varphi(X)=f(X)_c$; for a region $\setR$ the four corners of its $(M,G)$ cube are $X$ $(M{=}1,G{=}1)$, $\doop{\frz,\setR}X$ $(0,1)$, $\doop{\ablgeo,\setR}X$ $(1,0)$ and $\doop{\abl,\setR}X$ $(0,0)$, with outputs $\varphi$, $\varphi_{\frz}$, $\varphi_{\ablgeo}$, $\varphi_{\abl}$.

\paragraph{Proposition~\ref{prop:interact} (completeness, Shapley, interaction).}
Consider the two-player game on $\{M,G\}$ whose worth of a coalition $K$ is $v(K)=\varphi$ evaluated at the corner where exactly the factors in $K$ are \emph{kept}: $v(\{M,G\})=\varphi$, $v(\{G\})=\varphi_{\frz}$, $v(\{M\})=\varphi_{\ablgeo}$, $v(\emptyset)=\varphi_{\abl}$. The Shapley value of $M$ averages its marginal contribution over the two orders: $\tfrac12[(v(\{M,G\})-v(\{G\}))+(v(\{M\})-v(\emptyset))]=\tfrac12[(\varphi-\varphi_{\frz})+(\varphi_{\ablgeo}-\varphi_{\abl})]=\Smot$, and likewise for $G$, giving $\Sgeo$. Efficiency of the Shapley value gives $\Smot+\Sgeo=v(\{M,G\})-v(\emptyset)=\varphi-\varphi_{\abl}=\Nec$, for every $f$. The Shapley interaction index of the pair~\cite{grabisch1999interaction} is $v(\{M,G\})-v(\{M\})-v(\{G\})+v(\emptyset)=\Delta_{\mathrm{int}}$. The motion-first score $\varphi-\varphi_{\frz}$ equals $\Smot+\Delta_{\mathrm{int}}/2$ and the geometry-first one $\varphi_{\ablgeo}-\varphi_{\abl}$ equals $\Smot-\Delta_{\mathrm{int}}/2$ (direct substitution), so both deviate from $\Smot$ by $\pm\Delta_{\mathrm{int}}/2$. Finally, $\varphi$ restricted to the cube is additively separable, $\varphi(M,G)=a(M)+b(G)$, iff its mixed second difference vanishes, \ie iff $\Delta_{\mathrm{int}}=0$. $\square$

\paragraph{Proposition~\ref{prop:ace} (motion effects).}
By \textbf{(I)} the operators rewrite only $\setR$'s per-frame sets, so $\doop{\frz,\setR}$ sets $M_\setR{=}0$ while holding $G_\setR$ (the real shape, at its $t_0$ configuration) and all off-region content fixed: $\varphi-\varphi_{\frz}$ is the single-factor interventional contrast of $\setR$'s motion at its real geometry, a CaCE-type effect~\cite{goyal2019cace}; $\varphi_{\ablgeo}-\varphi_{\abl}$ is the same contrast for the shape-free proxy (same centroid path vs.\ static). Under \textbf{(S)}, $\varphi(M,G)=a(M)+b(G)$ on the cube and both contrasts equal $a(1)-a(0)$, the marginal ACE of motion; $\Smot$, their average, equals it too. The geometry statement is symmetric. $\square$

\paragraph{Proposition~\ref{prop:pnps} (necessity).}
Let $E=\mathbf 1[\varphi\ge\kappa]$. $E(X){=}1$ is Pearl's factual event and $E(\doop{\abl,\setR}X){=}0$ the counterfactual outcome under the do-operation that replaces $\setR$'s content (valid by \textbf{(I)}); substituting into the definition of $\PN$~\cite{pearl2009causality} gives $\PN(\setR)=\Pr[E(\doop{\abl,\setR}X){=}0\mid E(X){=}1]$. Under monotonicity and exogeneity \textbf{(M)}, Pearl's identification gives the excess-risk ratio $[P(E{=}1\mid\text{intact})-P(E{=}1\mid\text{replaced})]/P(E{=}1\mid\text{intact})$, whose numerator is the expectation of the hard-thresholded $\Nec$; the real-valued $\Nec$ is its margin relaxation, matching the local $\PN$ of~\cite{watson2021lens}. The sufficiency statement is dual. $\square$

\paragraph{Proposition~\ref{prop:context} (context-averaged decomposition).}
Completeness holds by construction: $\Smot+\Sgeo=\mu+\gamma+\iota=\Nec$ for every region. For the population statement, let $Z=(Z_1,\dots,Z_m)$ with $Z_j=(M_j,G_j)$ be the independent Bernoulli$(q)$ factors and $Y=\varphi(X_Z)$. Under independence $Y$ has the ANOVA (Hoeffding) decomposition $Y=\mu_0+\sum_j f_j(Z_j)+\sum_{j<k}f_{jk}(Z_j,Z_k)+\dots$, where every higher-order term has zero conditional mean given any single block $Z_j$, hence is $L^2$-orthogonal to every function of a single block. The regressors of Eq.~\eqref{eq:context}, $\{1\}\cup\{M_j,G_j,M_jG_j\}_j$, span exactly the functions $c+\sum_j g_j(Z_j)$ (a function of two binary variables is spanned by $1,M,G,MG$). The population least-squares solution is therefore the $L^2$ projection $\mu_0+\sum_j f_j(Z_j)$, with $f_j(z)=\Ex[Y\mid Z_j{=}z]-\Ex[Y]$. The fitted coefficients of block $j$ are thus the cube contrasts of $\bar\varphi_j(M,G)=\Ex[Y\mid M_j{=}M,G_j{=}G]$---the region's four corners averaged over random contexts of all other regions---and $\Smot=\mu+\iota/2$, $\Sgeo=\gamma+\iota/2$, $\Delta_{\mathrm{int}}=\iota$ are exactly Eqs.~\eqref{eq:sym}--\eqref{eq:dint} applied to $\bar\varphi_j$ (for $q=\tfrac12$: Banzhaf-type~\cite{banzhaf1965} main effects of the two factors, split symmetrically). If $\varphi$ has no cross-region interactions, every context gives the same contrasts, so $\bar\varphi_j$'s contrasts coincide with the local ones. $\square$

\section{Implementation details}
\label{supp:impl}
\paragraph{Data and skeleton ground truth.}
MSR-Action3D~\cite{li2010msraction3d} (567 depth videos of 20 actions by 10 subjects, 320$\times$240) is converted to point-cloud videos exactly as in MeteorNet~\cite{liu2019meteornet} (focal 280, principal point at the image centre, $z$ = depth; the input format of all three official backbones), then sampled to 24-frame centred clips of 2048 points per frame (1024 at laptop scale; independently per frame) and divided by 300. The dataset's 20-joint skeletons store $(u,v,d)$ per joint in depth-image pixels; we lift each joint with the same projection, taking its depth from the body surface around $(u,v)$ (median of the non-zero $7{\times}7$ window; joints outside the silhouette use a per-frame unit ratio fitted on the others). Over the \NumGTValid\ benchmark clips with a complete skeleton, a median \NumGTFitMedian\% of the points (10th percentile: \NumGTFitPten\%) lie within 25 units of a bone. One unit is about 5\,mm: the median head-to-foot joint distance is 318 units, which we take to be about 1.55\,m, so 25 units are about 12\,cm. Each point is assigned to its nearest of 19 bone segments; a bone is \emph{moving} if its midpoint path length over the clip is at least $0.5\times$ the clip's maximum and \emph{static} if at most $0.2\times$; moving-GT / static-GT are the points of moving / static bones (on average \NumGTMovingPct\% / \NumGTStaticPct\% of the points of a benchmark clip). \NumGTMissing\ benchmark clips, all from \emph{bend} and \emph{pick up \& throw}, store all-zero joints for a run of frames; their nearest-bone labels are meaningless there, so they are left out of every ground-truth-based number (they still count for faithfulness, which needs no ground truth).

\paragraph{Tracked regions and operators.}
Regions are tracked k-means supervoxels ($m{=}12$ unless stated): frame 0 uses farthest-point initialisation from the lowest point (so the partition does not depend on point order), each later frame re-runs k-means from the previous frame's centres (4 iterations). Operators act on the per-frame index sets: \frz\ fills the slots of $\setR_t$ with the points of $\setR_{t_0}$ (paired cyclically; $t_0$ = first frame); \ablgeo\ writes the centroid $c_t(\setR)$ into every slot of $\setR_t$ (a point mass on the centroid path); \abl\ writes $c_{t_0}(\setR)$; \dsp\ fills $\setR_t$ from $\setR_{\pi(t)}$. Because each operator's new values depend only on the region's own points, the joint state of all regions is a per-slot selection among three precomputed views, which lets both estimators evaluate hundreds of intervened clips per batch. A test in the released code checks that shuffling every frame's point order leaves the scores unchanged ($<10^{-6}$).

\paragraph{Backbones.}
No pretrained MSR weights exist for any of the three models, so we train the \emph{unmodified official model code} from scratch. Their compiled CUDA ops (FPS, ball query, grouping; \texttt{mamba\_ssm}; \texttt{knn\_cuda}) are replaced by numerically equivalent pure-PyTorch versions (FPS seeded at index 0, ball query returning the first in-radius indices, the reference selective scan), so the models also run on CPU/Apple GPUs. On CUDA, two of these loops are replaced by kernels that are checked against them on random inputs before use: a single-kernel FPS with the same selection rule, and the compiled selective scan of \texttt{mamba\_ssm}. \emph{Full scale} (main text): the official configuration of every model (44.1\,M / 8.3\,M / 109.8\,M parameters for P4Transformer / PSTNet / Mamba4D), 2048 points per frame, one A100 GPU; AdamW ($3{\times}10^{-4}$, weight decay 0.05), 5 warm-up epochs then cosine decay, the official batch sizes (14 / 16 / 16), 50 / 35 / 50 epochs, 17 random 24-frame clips per training video per epoch (about the number the official loaders enumerate), the official augmentation (random per-axis scaling in $[0.9,1.1]$), TF32 matrix products during training and FP32 for explanation. \emph{Laptop scale} (Appendix~\ref{supp:scale}): P4Transformer dim 256, depth 4, 4 heads of 64 (spatial stride 16 for 1024 points, \ie the official 64 anchors per frame), Mamba4D dim 256 with 2 inter- and 1 intra-frame Mamba layers, PSTNet official; 1024 points, batch 8, 50 / 30 / 20 epochs, 5 / 3 / 3 clips per video per epoch, on an Apple M4. Both scales report the last epoch; no model is selected on the test set. Two details were necessary for from-scratch training to leave its initial plateau: P4Transformer and Mamba4D embed \emph{absolute} $(x,y,z,t)$ with $t=1,\dots,L$, so at initialisation the time channel dominates every token ($\|\text{pos}\|\approx34$ vs.\ $\|\text{feature}\|\approx7$) and the token max-pool selects tokens by frame index rather than content; we therefore scale the \emph{initial} weights of that channel by $1/L$ (the architecture is unchanged) and subtract each clip's centroid (a preprocessing step applied identically at explanation time). With the official SGD recipe the reduced models did not leave the plateau within the $\approx$10$^3$ iterations a laptop affords, so both scales use AdamW.

\paragraph{Baselines and metrics.}
Integrated Gradients and Extremal Perturbations use the clip centroid as their reference (a far-away constant is off-manifold for real point networks); 3D Grad-CAM uses the gradient-weighted tube-embedding tokens of P4Transformer/Mamba4D, scattered to points through their nearest anchors (an input-space proxy for PSTNet); attention rollout and transformer attribution (generic attention explainability~\cite{chefer2021transformer}) use all P4Transformer layers; raw attention is the attention each token receives in the last layer, scattered to points; region occlusion deletes each tracked region; OccAM averages the output over 128 random 50\% point masks; Extremal optimises a per-point mask for 30 Adam steps at 10\% area. Deletion/insertion remove/insert points in importance order in 20 steps; a removed point is deleted and its frame re-sampled from the remaining points (the official loaders' rule for short frames); all curves score the target-class \emph{probability}. Comprehensiveness/Sufficiency use AOPC over $\{1,5,10,20,50\}\%$; Sensitivity-$n$ uses 30 random 10\% subsets.

\section{Controlled experiment: ground truth by construction}
\label{supp:controlled}
To test the decomposition where the answer is known \emph{by construction}, every clip has three point clusters with the same layout in all classes; one ``odd'' cluster differs and the class is \emph{which} cluster is odd. In the \textbf{motion} task the odd cluster translates (all shapes identical: only motion is discriminative); in the \textbf{shape} task it is larger and nothing moves (only geometry is discriminative). An attention-based 4D classifier (per-point MLP $\to$ per-frame voxel tokens $\to$ Transformer) is trained per task and seed (3 seeds, 20 held-out clips each) and explained with tracked regions ($m{=}3$). Within the odd cluster, the motion share of \method's positive importance is \NumCtrlMot\ for the motion model and \NumCtrlShape\ for the shape model (local estimator; context: \NumCtrlMotCtx\ vs.\ \NumCtrlShapeCtx), \ie a complete double dissociation. With the earlier \emph{deletion} corner instead of the static proxy, a static region's presence leaks into $\Smot$ (Tab.~\ref{tab:controlled-abl}). On a third task whose class is the motion \emph{direction} and whose two directions share the same spatial envelope, reversing the actor's trajectory (\dsp) flips the prediction to the opposite class in \NumCtrlDspFlip\% of the clips, freezing it in \NumCtrlFrzFlip\%: \dsp\ is sign-sensitive, \frz\ is not.

\begin{table}[h]
\centering\small\tablecitesoff
\begin{tabular}{l ccc}
\toprule
$(0,0)$ corner, order & mot. & shape & Del.$\downarrow$\\
\midrule
static proxy (default), symmetric & 1.00 & 0.00 & 0.428\\
static proxy, motion-first & 1.00 & 0.00 & 0.428\\
deletion, symmetric & 0.92 & 0.49 & 0.419\\
deletion, motion-first & 0.85 & 0.00 & 0.419\\
donor fill, symmetric & 0.92 & 0.49 & 0.419\\
donor fill, motion-first & 0.85 & 0.00 & 0.419\\
\bottomrule
\end{tabular}

\caption{Controlled ablation (1 seed): the $(0,0)$ corner and the order. Motion share within the odd cluster on the motion (mot.) and shape (shape) models; the decomposition is correct iff mot.\ $\approx1$ and shape $\approx0$.}
\label{tab:controlled-abl}
\end{table}

\section{Data, models and code behind each result}
\label{supp:provenance}
\paragraph{Data.} MSR-Action3D~\cite{li2010msraction3d}: raw depth maps and 20-joint skeletons from the archive distributed with MeteorNet~\cite{liu2019meteornet}, converted with MeteorNet's depth-to-point procedure (Appendix~\ref{supp:impl}). Every real-data result uses test subjects 6--10 only; models are trained on subjects 1--5. Each clip is the centred 24-frame window of a video, and its points are sampled with a fixed seed per video, so every method explains exactly the same inputs.

\paragraph{Models and code.} The unmodified official repositories of P4Transformer~\cite{fan2021p4transformer} (\url{github.com/hehefan/P4Transformer}, commit \texttt{461b83e}), PSTNet~\cite{fan2021pstnet} (\url{github.com/hehefan/Point-Spatio-Temporal-Convolution}, \texttt{04f9b47}) and Mamba4D~\cite{liu2025mamba4d} (commit \texttt{8d5bbe2}). Their compiled operators are replaced as described in Appendix~\ref{supp:impl}; on CUDA the selective scan is the kernel of \texttt{mamba\_ssm} 2.3.2~\cite{gu2024mamba}. Full scale: PyTorch~\cite{paszke2019pytorch} 2.11 with CUDA 13.0 on one A100 (40\,GB). Laptop scale: PyTorch 2.14 on an Apple M4. Statistical tests use SciPy~\cite{virtanen2020scipy}.

\paragraph{From records to numbers.} Every explained clip is written to a per-clip record (prediction, attribution scores, every metric). Scripts compute every table body and every measured number in the text from these records; no measured number is typed by hand. Tab.~\ref{tab:provenance} lists the record behind each result.

\begin{table*}[h]
\centering\small\tablecitesoff
\begin{tabular}{p{0.22\textwidth} p{0.31\textwidth} p{0.14\textwidth} p{0.20\textwidth}}
\toprule
Result & Data & Model(s) & Record\\
\midrule
Tab.~\ref{tab:faithfulness}, \ref{tab:mpg} (left), \ref{tab:faith-supp}, \ref{tab:cost}, \ref{tab:per-class}; Fig.~\ref{fig:curves}, \ref{fig:teaser}, \ref{fig:qual} & all \NumPFTClips\ test clips, 2048 points per frame; ground-truth metrics on the \NumGTValid\ with a complete skeleton & P4Transformer (Fig.~\ref{fig:qual}: all three) & benchmark, \url{p4transformer/raw.jsonl}\\
Tab.~\ref{tab:cross} & the first two test clips per action (\NumCrossClips) & all three & benchmark, one file per backbone\\
Tab.~\ref{tab:mpg} (right), Fig.~\ref{fig:dissoc}, Appendix~\ref{supp:results} & arm-moving videos: \NumXTrain\ train, \NumXTest\ test; \NumXEval\ test clips explained per task & one task model per task and backbone & \url{crossover*/crossover.json}\\
Tab.~\ref{tab:ablation}, Fig.~\ref{fig:sanity}; displace probe & \NumCrossClips\ clips (sanity: the first 20) & P4Transformer & \url{ablation/ablation.json}\\
Tab.~\ref{tab:stability} & the same \NumCrossClips\ clips, \NumStabSamplings\ point samplings each & P4Transformer & \url{stability/stability.jsonl}\\
Tab.~\ref{tab:seeds} & as the crossover, \NumXSeedsTotal\ training seeds & P4Transformer task models & \url{crossover*/crossover.json}\\
Tab.~\ref{tab:controlled-abl}; controlled bar of Fig.~\ref{fig:dissoc} & synthetic clips, 3 seeds $\times$ 20 test clips & small attention model & \url{controlled/*.json}\\
Tab.~\ref{tab:scale} & as above, laptop-scale models & all three & laptop-scale records\\
\bottomrule
\end{tabular}
\caption{The data, models and per-clip records behind each table and figure. All real-data results use the MSR-Action3D test subjects 6--10 and, except Tab.~\ref{tab:scale}, the full-scale models.}
\label{tab:provenance}
\end{table*}

\begin{table*}[h]
\centering\small\tablecitesoff
\begin{tabular}{l cc cc}
\toprule
& \multicolumn{2}{c}{one sampling} & \multicolumn{2}{c}{average of two samplings}\\
\cmidrule(lr){2-3}\cmidrule(lr){4-5}
Method & points & body parts & points & body parts\\
\midrule
Random (floor) & 0.00 & $-$0.01 & 0.00 & 0.01\\
Motion-prior (floor) & 0.66 & 0.99 & 0.73 & 0.99\\
Grad$\times$Input & 0.82 & 0.86 & 0.81 & 0.87\\
Integrated Grad.~\cite{sundararajan2017ig} & 0.92 & 0.94 & 0.91 & 0.93\\
Region occlusion~\cite{zheng2019pointcloudsaliency} & 0.71 & 0.85 & 0.81 & 0.91\\
OccAM~\cite{schinagl2022occam} & 0.00 & 0.04 & 0.00 & 0.04\\
Attn.\ Rollout~\cite{abnar2020rollout} & 0.68 & 0.70 & 0.75 & 0.76\\
\midrule
\textbf{\method} (local) & 0.43 & 0.68 & 0.57 & 0.80\\
\textbf{\method} (context) & 0.62 & 0.82 & 0.76 & 0.89\\
\bottomrule
\end{tabular}
\caption{\textbf{Stability under sensor re-sampling} (P4Transformer, \NumStabClips\ test clips, \NumStabSamplings\ independent point samplings each). Spearman rank correlation between the map of sampling 0 and the map of another sampling transferred onto the points of sampling 0 (nearest neighbour per frame), averaged over samplings and clips. Body parts: each map averaged over the points of each of the 19 skeleton bones (one clip without a complete skeleton is left out). Average of two samplings: maps of samplings 0 and 1 averaged and compared with the average of samplings 2 and 3. Method randomness is fixed per clip, so only the point sampling changes.}
\label{tab:stability}
\end{table*}

\section{Two training scales}
\label{supp:scale}
The main text uses backbones trained at their official width with 2048 points per frame on one A100 GPU. We also trained all three at laptop scale (reduced width for P4Transformer and Mamba4D, 1024 points, an Apple M4; Appendix~\ref{supp:impl}) and ran every real-data experiment again. Tab.~\ref{tab:scale} sets the headline numbers side by side. The full-scale models are closer to their published accuracies, most of all Mamba4D. The conclusions agree, with three differences in significance. On P4Transformer, \method's deletion advantage over region occlusion is significant at laptop scale on \NumPFTClipsFive\ clips and at full scale on all \NumPFTClips\ clips ($p\NumPKcaCtxOcclDel$), but not at full scale on the same \NumPFTClipsFive\ clips ($p\NumPKcaCtxOcclDelFive$): the full test set is what makes it significant. On Mamba4D its advantage over Integrated Gradients is significant only at laptop scale. Integrated Gradients localises the moving parts better than $\Smot$ on P4Transformer at full scale, where the laptop-scale run had them tied. In the crossover the shape-task models of P4Transformer and Mamba4D draw more of the arm's importance from motion at full scale, but every motion--shape gap still excludes zero.

\begin{table*}[h]
\centering\small\tablecitesoff
\begin{tabular}{l cc}
\toprule
Quantity & laptop & full scale\\
\midrule
Test accuracy, P4Transformer (\%) & 90.6 & 89.2\\
Test accuracy, PSTNet (\%) & 87.5 & 88.9\\
Test accuracy, Mamba4D (\%) & 83.6 & 88.2\\
\midrule
P4T Del.: \method\ (context) & 0.178 & 0.179\\
P4T Del.: region occlusion & 0.219 & 0.233\\
P4T Del.: Integrated Gradients & 0.268 & 0.282\\
P4T Del.: context vs.\ occlusion & $p{=}0.0043$ & $p{<}10^{-3}$\\
P4T Del.: context vs.\ IG & $p{<}10^{-3}$ & $p{<}10^{-3}$\\
PSTNet Del.: context / occlusion & 0.191 / 0.222 & 0.217 / 0.252\\
Mamba4D Del.: context / occlusion & 0.245 / 0.240 & 0.269 / 0.246\\
Mamba4D Del.: context vs.\ IG & $p{<}10^{-3}$ & $p{=}0.064$\\
P4T $\Smot$ EBPG: context / IG & 0.54 / 0.58 & 0.50 / 0.55\\
\midrule
Crossover P4T: motion / shape task & 0.81 / 0.39 & 0.83 / 0.60\\
Crossover PSTNet: motion / shape task & 0.85 / 0.23 & 0.86 / 0.30\\
Crossover Mamba4D: motion / shape task & 0.89 / 0.54 & 0.92 / 0.61\\
\midrule
Randomised model: \method\ / Grad$\times$Input & $-$0.01 / 0.72 & $-$0.10 / 0.81\\
\bottomrule
\end{tabular}
\caption{Headline numbers at the two training scales (laptop: \NumPFTClipsFive\ P4Transformer clips; full scale: all \NumPFTClips; the other rows use the same clips at both scales. Del.: deletion AUC; crossover: motion share of the arm's importance; randomised model: rank correlation with the trained model's map once every block is re-initialised).}
\label{tab:scale}
\end{table*}

\section{Additional real-data results}
\label{supp:results}
\paragraph{Crossover design.} We explain, in each task, the class whose defining property is \emph{present} in the clip and removable by the interventions: the original class on original clips in the motion task (its normal-speed motion is removed by \frz) and the thick class on thickened clips in the shape task (its shape is removed by the proxy). We report the variants we tried on the way. Explaining the \emph{original} class in the shape task---whose evidence is then the \emph{absence} of the cue---gave shares that depend on the sign convention (positive parts \NumXThickShare, signed sums \NumXThickShareSigned): freezing a normally shaped arm moves that logit strongly in both directions from clip to clip. Thinning the arm instead of thickening it was not learnable by P4Transformer (test accuracy \NumXThinAcc\%, i.e.\ chance), so that variant gives no information. With the final design both conventions agree (motion task: \NumXMotShare\ positive / \NumXMotShareSigned\ signed; shape task: \NumXGeoShare\ / \NumXGeoShareSigned). 95\% bootstrap intervals over clips (positive parts, local estimator): motion task [\NumXMotLo, \NumXMotHi], shape task [\NumXGeoLo, \NumXGeoHi], gap [\NumXGapLo, \NumXGapHi]. We retrained both task models with further seeds (Tab.~\ref{tab:seeds}). At full scale the motion-task model often fails to leave chance: it trains in \NumXSeedsTrained\ of \NumXSeedsTotal\ seeds, and the shares of a model at chance carry no information, so we do not use them. In all \NumXSeedsSameDir\ seeds whose models trained, the arm's motion share is higher in the motion task; the interval of the gap excludes zero in \NumXSeedsCIExcl\ of them, and the smallest gap (seed 3) comes with the weakest shape model. Three of the crossover videos (one training, two test, none of them among the explained clips) have frames without a skeleton; their arm masks are wrong in those frames, and we keep them.

\begin{table*}[h]
\centering\small\tablecitesoff
\begin{tabular}{l ccc}
\toprule
Seed & acc.\ \% (mot.\ / shape) & share (mot.\ / shape) & gap 95\% CI\\
\midrule
0 & 80.5 / 83.5 & 0.83 / 0.60 & [0.14, 0.32]\\
1 \emph{(not trained)} & 50.0 / 86.5 & 0.65 / 0.55 & [$-$0.02, 0.21]\\
2 & 67.8 / 88.8 & 0.73 / 0.57 & [0.06, 0.26]\\
3 & 78.8 / 77.5 & 0.85 / 0.78 & [$-$0.02, 0.19]\\
4 \emph{(not trained)} & 50.0 / 72.2 & 0.60 / 0.81 & [$-$0.32, $-$0.08]\\
5 \emph{(not trained)} & 50.0 / 80.5 & 0.73 / 0.69 & [$-$0.05, 0.17]\\
\bottomrule
\end{tabular}
\caption{P4Transformer crossover over training seeds (local estimator, \NumXEval\ explained test clips per task). A seed counts only if both task models leave chance (accuracy $\geq 60\%$ on the balanced binary test set); seed 0 is the one reported in the main text.}
\label{tab:seeds}
\end{table*}
 \emph{PSTNet} (same tasks and recipe): \NumXPstMotShare\ [\NumXPstMotLo, \NumXPstMotHi] and \NumXPstGeoShare\ [\NumXPstGeoLo, \NumXPstGeoHi] (context estimator \NumXPstMotShareCtx\ / \NumXPstGeoShareCtx; motion prior on the arm \NumXPstMotPriorEBPG\ / \NumXPstGeoPriorEBPG\ EBPG in both tasks); the signed shares (\NumXPstMotShareSigned, \NumXPstGeoShareSigned) agree in direction and fall just outside $[0,1]$ because the other term's sum over the arm is slightly negative. \emph{Mamba4D}: \NumXMbaMotShare\ [\NumXMbaMotLo, \NumXMbaMotHi] and \NumXMbaGeoShare\ [\NumXMbaGeoLo, \NumXMbaGeoHi] (context \NumXMbaMotShareCtx\ / \NumXMbaGeoShareCtx; signed \NumXMbaMotShareSigned\ / \NumXMbaGeoShareSigned).
\begin{table*}[h]
\centering\small\tablecitesoff
\begin{tabular}{l cccccc}
\toprule
Variant & Del.$\downarrow$ & $\Smot$ EBPG$\uparrow$ & mot.\ share & $\tfrac{|\Delta_{\mathrm{int}}|}{|\Nec|}$ & passes & s\\
\midrule
local, $m{=}6$ & 0.235 & 0.47 & 0.65 & 0.46 & 19 & 0.3\\
local, $m{=}12$ (default) & 0.272 & 0.54 & 0.73 & 0.63 & 37 & 0.5\\
local, $m{=}24$ & 0.334 & 0.54 & 0.78 & 0.70 & 73 & 0.8\\
\midrule
context, $m{=}6$ & 0.192 & 0.44 & 0.65 & 0.48 & 257 & 2.3\\
context, $m{=}12$ (default) & 0.219 & 0.54 & 0.75 & 0.58 & 257 & 2.3\\
context, $m{=}24$ & 0.212 & 0.52 & 0.82 & 0.66 & 257 & 2.4\\
\midrule
local, removal instead of proxy & 0.235 & 0.49 & 0.70 & 0.75 & 37 & 0.5\\
context, removal instead of proxy & 0.126 & 0.51 & 0.72 & 0.58 & 257 & 3.0\\
local, donor fill instead of proxy & 0.208 & 0.50 & 0.70 & 0.72 & 37 & 2.1\\
local, motion-first order & 0.272 & 0.53 & 0.88 & -- & 25 & 1.5\\
context, $S{=}64$ & 0.231 & 0.49 & 0.75 & 0.77 & 65 & 0.7\\
context, $S{=}128$ & 0.204 & 0.53 & 0.76 & 0.64 & 129 & 1.2\\
context, $S{=}512$ & 0.211 & 0.55 & 0.76 & 0.52 & 513 & 4.5\\
context, $q{=}0.75$ & 0.233 & 0.55 & 0.74 & 0.63 & 257 & 2.3\\
local, anchor $t_0$ = middle frame & 0.245 & 0.56 & 0.68 & 0.65 & 37 & 0.5\\
context, anchor $t_0$ = middle frame & 0.173 & 0.57 & 0.69 & 0.58 & 257 & 2.3\\
\bottomrule
\end{tabular}
\caption{\textbf{Ablation and cost} (P4Transformer, 40 clips). EBPG and motion share on the moving ground truth; $|\Delta_{\mathrm{int}}|/|\Nec|$: mean interaction relative to necessity; forward passes and seconds per clip on one A100 GPU.}
\label{tab:ablation}
\end{table*}

\begin{table*}[h]
\centering\small\tablecitesoff
\begin{tabular}{l ccccc}
\toprule
Method & Del.$\downarrow$ & Comp.$\uparrow$ & Ins.$\uparrow^\dagger$ & Suff.$\downarrow^\dagger$ & Sens-$n\uparrow$\\
\midrule
Random (floor) & 0.916{\scriptsize$\pm$0.14} & 0.003{\scriptsize$\pm$0.05} & 0.920{\scriptsize$\pm$0.14} & 0.041{\scriptsize$\pm$0.12} & 0.004{\scriptsize$\pm$0.17}\\
Geometry-edge (floor) & 0.890{\scriptsize$\pm$0.19} & 0.011{\scriptsize$\pm$0.11} & 0.903{\scriptsize$\pm$0.16} & 0.103{\scriptsize$\pm$0.21} & 0.000{\scriptsize$\pm$0.17}\\
Motion-prior (floor) & 0.505{\scriptsize$\pm$0.38} & 0.277{\scriptsize$\pm$0.33} & 0.860{\scriptsize$\pm$0.16} & 0.365{\scriptsize$\pm$0.22} & 0.000{\scriptsize$\pm$0.18}\\
Grad$\times$Input & 0.655{\scriptsize$\pm$0.28} & 0.086{\scriptsize$\pm$0.18} & 0.722{\scriptsize$\pm$0.25} & 0.332{\scriptsize$\pm$0.28} & \underline{0.028{\scriptsize$\pm$0.23}}\\
SmoothGrad~\cite{smilkov2017smoothgrad} & 0.772{\scriptsize$\pm$0.24} & 0.031{\scriptsize$\pm$0.12} & \underline{0.915{\scriptsize$\pm$0.14}} & \underline{0.083{\scriptsize$\pm$0.16}} & 0.006{\scriptsize$\pm$0.18}\\
Integrated Grad.~\cite{sundararajan2017ig} & 0.282{\scriptsize$\pm$0.26} & 0.347{\scriptsize$\pm$0.24} & 0.862{\scriptsize$\pm$0.17} & 0.129{\scriptsize$\pm$0.19} & \textbf{0.046{\scriptsize$\pm$0.22}}\\
3D Grad-CAM~\cite{selvaraju2017gradcam} & 0.713{\scriptsize$\pm$0.26} & 0.051{\scriptsize$\pm$0.12} & 0.554{\scriptsize$\pm$0.30} & 0.659{\scriptsize$\pm$0.30} & 0.000{\scriptsize$\pm$0.15}\\
Region occlusion~\cite{zheng2019pointcloudsaliency} & \underline{0.233{\scriptsize$\pm$0.23}} & \underline{0.416{\scriptsize$\pm$0.23}} & 0.784{\scriptsize$\pm$0.17} & 0.531{\scriptsize$\pm$0.24} & 0.021{\scriptsize$\pm$0.16}\\
OccAM~\cite{schinagl2022occam} & 0.853{\scriptsize$\pm$0.24} & 0.059{\scriptsize$\pm$0.13} & \textbf{0.953{\scriptsize$\pm$0.07}} & \textbf{$-$0.004{\scriptsize$\pm$0.11}} & 0.003{\scriptsize$\pm$0.17}\\
Extremal Pert.~\cite{fong2019extremal} & 0.634{\scriptsize$\pm$0.22} & 0.052{\scriptsize$\pm$0.10} & 0.309{\scriptsize$\pm$0.26} & 0.830{\scriptsize$\pm$0.25} & 0.005{\scriptsize$\pm$0.17}\\
Attn.\ Rollout~\cite{abnar2020rollout} & 0.649{\scriptsize$\pm$0.27} & 0.073{\scriptsize$\pm$0.15} & 0.769{\scriptsize$\pm$0.24} & 0.431{\scriptsize$\pm$0.26} & $-$0.009{\scriptsize$\pm$0.17}\\
Transformer Attr.~\cite{chefer2021transformer} & 0.601{\scriptsize$\pm$0.28} & 0.100{\scriptsize$\pm$0.17} & 0.766{\scriptsize$\pm$0.24} & 0.424{\scriptsize$\pm$0.28} & 0.003{\scriptsize$\pm$0.17}\\
Raw attention (last layer) & 0.542{\scriptsize$\pm$0.28} & 0.127{\scriptsize$\pm$0.16} & 0.688{\scriptsize$\pm$0.25} & 0.551{\scriptsize$\pm$0.29} & 0.006{\scriptsize$\pm$0.16}\\
\midrule
\textbf{\method} (local) & 0.256{\scriptsize$\pm$0.25} & 0.403{\scriptsize$\pm$0.24} & 0.793{\scriptsize$\pm$0.17} & 0.511{\scriptsize$\pm$0.23} & 0.018{\scriptsize$\pm$0.17}\\
\textbf{\method} (context) & \textbf{0.179{\scriptsize$\pm$0.19}}$^\ast$ & \textbf{0.453{\scriptsize$\pm$0.21}}$^\ast$ & 0.761{\scriptsize$\pm$0.18} & 0.546{\scriptsize$\pm$0.24} & 0.011{\scriptsize$\pm$0.17}\\
\bottomrule
\end{tabular}
\caption{Full faithfulness table on P4Transformer, including Sensitivity-$n$ (Pearson correlation between summed attribution and output drop over random 10\% subsets; near zero for all methods, because random subsets barely change the output).}
\label{tab:faith-supp}
\end{table*}

\begin{table}[h]
\centering\small\tablecitesoff
\begin{tabular}{l c}
\toprule
Method & s / clip\\
\midrule
Random (floor) & 0.00\\
Geometry-edge (floor) & 0.14\\
Motion-prior (floor) & 0.16\\
Grad$\times$Input & 0.06\\
SmoothGrad~\cite{smilkov2017smoothgrad} & 1.53\\
Integrated Grad.~\cite{sundararajan2017ig} & 1.93\\
3D Grad-CAM~\cite{selvaraju2017gradcam} & 0.07\\
Region occlusion~\cite{zheng2019pointcloudsaliency} & 0.57\\
OccAM~\cite{schinagl2022occam} & 1.46\\
Extremal Pert.~\cite{fong2019extremal} & 2.02\\
Attn.\ Rollout~\cite{abnar2020rollout} & 0.05\\
Transformer Attr.~\cite{chefer2021transformer} & 0.07\\
Raw attention (last layer) & 0.05\\
\textbf{\method} (local) & 0.47\\
\textbf{\method} (context) & 2.33\\
\bottomrule
\end{tabular}
\caption{Explanation cost per clip on one A100 GPU (P4Transformer, 24 frames $\times$ 2048 points; seconds, excluding metric evaluation).}
\label{tab:cost}
\end{table}

\begin{figure}[h]
\centering
\includegraphics[width=0.85\linewidth]{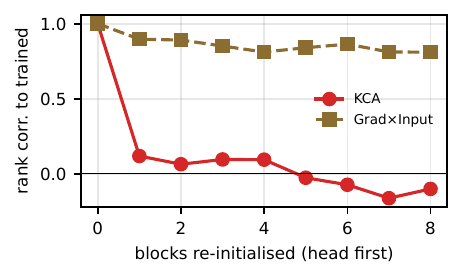}
\caption{Cascading model randomisation~\cite{adebayo2018sanity} of P4Transformer, from the classification head down: rank correlation of the attribution map with the trained model's.}
\label{fig:sanity}
\end{figure}

\begin{figure*}[p]
\centering
\includegraphics[width=\textwidth,height=0.92\textheight,keepaspectratio]{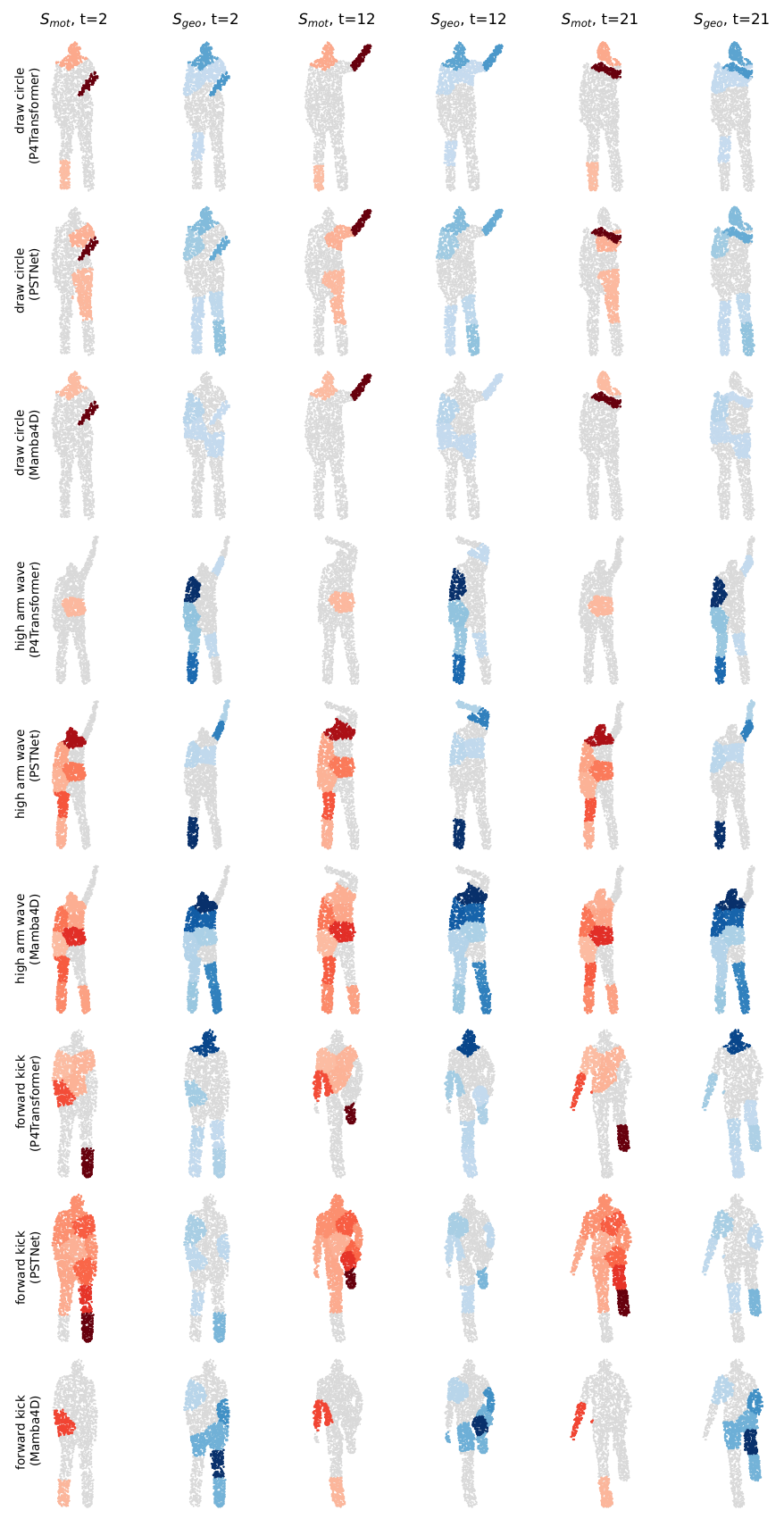}
\caption{$\Smot$ (red) and $\Sgeo$ (blue) on real MSR-Action3D test clips at three instants; rows: three actions (\emph{draw circle}, \emph{high arm wave}, \emph{forward kick}) $\times$ the three backbones, same code.}
\label{fig:qual}
\end{figure*}

\begin{table}[h]
\centering\small\tablecitesoff\setlength{\tabcolsep}{3pt}
\begin{tabular}{l cccc}
\toprule
Action & mot.\ share & Del.\ \method & Del.\ occl. & acc.\,\%\\
\midrule
high arm wave & 0.61 & 0.151 & 0.265 & 100\\
horizontal arm wave & 0.61 & 0.062 & 0.090 & 100\\
hammer & 0.63 & 0.074 & 0.075 & 7\\
hand catch & 0.68 & 0.244 & 0.359 & 80\\
forward punch & 0.63 & 0.193 & 0.232 & 92\\
high throw & 0.65 & 0.103 & 0.101 & 86\\
draw x & 0.67 & 0.060 & 0.061 & 77\\
draw tick & 0.72 & 0.085 & 0.144 & 93\\
draw circle & 0.64 & 0.042 & 0.045 & 80\\
hand clap & 0.73 & 0.071 & 0.084 & 100\\
two hand wave & 0.66 & 0.080 & 0.078 & 100\\
side-boxing & 0.69 & 0.280 & 0.310 & 71\\
bend & 0.73 & 0.325 & 0.291 & 93\\
forward kick & 0.80 & 0.494 & 0.698 & 100\\
side kick & 0.71 & 0.209 & 0.437 & 100\\
jogging & 0.54 & 0.422 & 0.626 & 100\\
tennis swing & 0.74 & 0.078 & 0.097 & 73\\
tennis serve & 0.75 & 0.163 & 0.176 & 87\\
golf swing & 0.61 & 0.261 & 0.223 & 100\\
pick up \& throw & 0.69 & 0.231 & 0.331 & 100\\
\bottomrule
\end{tabular}
\caption{Per action (P4Transformer, all test clips of each action): motion share $\Smot^+/(\Smot^++\Sgeo^+)$ of \method\ (context) over the whole body, deletion AUC of \method\ (context) and of region occlusion, and the model's accuracy on these clips.}
\label{tab:per-class}
\end{table}

\end{document}